\pdfoutput=1

\PassOptionsToPackage{table, prologue}{xcolor}

\documentclass[11pt]{article}

\usepackage[]{ACL2023}

\usepackage{times}
\usepackage{latexsym}

\usepackage[T1]{fontenc}

\usepackage[utf8]{inputenc}

\usepackage{microtype}

\usepackage{inconsolata}

\usepackage{xspace}
\usepackage{booktabs}
\usepackage{multicol}
\usepackage{amsmath}
\usepackage{multirow}
\usepackage{graphicx}
\usepackage{enumitem}
\usepackage{mdwlist}
\usepackage{textcomp}
\usepackage{siunitx}
\usepackage{longtable}
\usepackage{verbatim}

\newcommand{\camerareadytext}[1]{#1\xspace}
\newcommand{\cuttext}[1]{}

\newcommand{\objexp}{\emph{Objective Experience}}
\newcommand{\advice}{\emph{Advice}}

\newcommand{\dataset}{\textsc{AloE}\xspace}

\newcommand{\myparagraph}[1]{\noindent \textbf{#1}}

\newcommand{\fref}[1]{Figure~\ref{#1}}
\newcommand{\tref}[1]{Table~\ref{#1}}

\title{Modeling Empathetic Alignment in Conversation}

\author{Jiamin Yang \\
        University of Chicago \\
       \texttt{jiaminy@uchicago.edu} \\\And
        David Jurgens \\
        University of Michigan \\
        \texttt{jurgens@umich.edu} \\}

\begin{document}
\maketitle
\begin{abstract}

Empathy requires perspective-taking: empathetic responses require a person to reason about what another has experienced and communicate that understanding in language. 
However, most NLP approaches to empathy do not explicitly model this alignment process.
Here, we introduce a new approach to recognizing alignment in empathetic speech, grounded in Appraisal Theory. 
We introduce a new dataset of over 9.2K span-level annotations of different types of appraisals of a person's experience and over 3K empathetic alignments between a speaker's and observer's speech.
Through computational experiments, we show that these appraisals and alignments can be accurately recognized.
In experiments in over 9.2M Reddit conversations, we find that appraisals capture meaningful groupings of behavior but that most responses have minimal alignment. However, we find that mental health professionals engage with substantially more empathetic alignment.

\end{abstract}

\section{Introduction}

Empathy is a key aspect of successful clinical health conversations \citep{hojat2013empathy,raab2014mindfulness}. In general,  empathy involves an emotional component, where a listener resonates with the emotional tone of a speaker, and a cognitive component, conveying the listener understands the speaker \cite{hatfield2011emotional}. Underlying both of these components is the perspective-taking by the listener to mirror the experience of the speaker, or as \citet{Mahrer1997EmpathyAT} describes it, ``being aligned is another way of being empathic." While past computational work on empathy has measured how empathetic messages can be, we still understand little about what aligns the language and perspective. Here, we examine empathy as an alignment task, studying therapeutic conversations on Reddit.

Given the importance of empathy, particularly in the clinical setting, NLP methods have attempted to model the relative level of empathy in replies \citep{sharma2020computational,omitaomu2022empathic}. Better models for recognizing empathy are aimed to help support generating more empathetic responses \citep[e.g.,][]{sharma2021facilitating,welivita2023empathetic}. 
However, as \citet{lahnala2022critical} note, many of these works focus only on the emotional mirroring component of empathy, rather than its cognitive component of perspective taking, and none explicitly model the alignment between the speaker, known as the \textit{Target}, and listener, as the \textit{Observer}.

Here, we introduce a new dataset and computational models for studying empathetic alignment in conversation. To quantify alignment, our work draws on the Appraisal Theory \citep{appraisalWondra}, which describes six aspects of how a person may experience a situation, e.g., describing its pleasantness or how much control they had, and encompasses both cognitive and emotional components. This scheme gives us a fine-grain labeling of both what is described and how the person feels. Because both the Target and Observer can appraise the same content differently, this view provides critical insight for understanding whether the two are aligned. 

This paper offers the following four contributions to the study of empathy in NLP. 
First, we introduce \dataset, a new dataset, of therapeutic Reddit conversations labeled with 9,284 appraisals from both the Target and Observer and 3,262 alignments between the Target and Observer.  Our dataset goes beyond theory to introduce new categories that model common types of aligned spans.
Second, in experiments, we show that appraisals can be accurately recognized and that the alignment between appraisals can be recognized, though we show that both are challenging tasks.
Third, in analyses on the appraisals and alignments of 2.3M posts and 8.9M comments, we show that appraisals meaningfully capture differences in how individuals experience distressing situations and in how others reply---but that the dominant form of alignment is to reply with advice, rather than a matched appraisal.
Fourth, in comparisons between mental health professionals and laypeople on Reddit, professionals have much higher alignment with Targets; but, as seen in clinical settings, both professionals and laypeople decrease in their levels of alignment as they become more experienced.

\section{Empathy in Therapeutic Settings}

Empathy has been an important concept in social, personality, and clinical psychology \cite{davis2018empathy, eisenberg2013prosocial, batson1981empathic, hall_clinical}. Though being diversely defined, the most discussed aspects are emotional empathy and cognitive empathy \cite{cuff_empathy_review}. Emotional empathy focuses on the vicarious sharing of emotion, while cognitive empathy relates to mental perspective-taking \cite{smith2006cognitive, shamay2011neural, BLAIR2005698}. In other words, emotional empathy is expressed as "I feel what you feel", and cognitive empathy is more commonly recognized as "I understand what you feel" \cite{healey2018cognitive}. 

Empathetic conversation is thought to play an important role in the development of social relationships \cite{hoffman2001empathy}, and mental health professionals are taught to develop empathetic skills \cite{toombs2001role, moudatsou2020role}, to improve patient outcomes\camerareadytext{ and experiences}. 
Central to these empathetic conversations is the alignment between what a Target is feeling and confirmation that the Observer's mental model of the Target matches these feelings; explicit expressions of this alignment are important for a Target to experience an Observer's response as empathetic \citep[e.g.,][]{thwaites2007conceptualizing,vyskocilova2011empathy,watson2016role}. 
While related to concepts like ``active listening'' or ``reflective listening,'' this type of speech requires a communication of the Observer’s theory of mind to show that they have understood what the Target has experienced, rather than just repeating parts of what a Target has said.

Individuals seeking mental health support increasingly turn to social media  \cite{hanley2019systematic}. Compared with traditional therapy sessions, the observers are no longer guaranteed to be trained professionals and the interactions are largely text-only. Given abundant data and unique features, empathy in online communities becomes a valuable subject for active research \cite{naslund2016future}, including comparisons of defining and expressing empathy between laypeople and professionals  \cite{laypeopleEmpathy, lahnala2021exploring}.

Within NLP, significant work has been done in predicting empathy \cite{guda2021empathbert,vasava2022transformer}, analyzing empathetic expressions and behaviors \cite{sharma2020computational,zhou2020condolence}, and facilitating empathetic conversations \cite{sharma2021facilitating, xie2021empathetic, zeng2021affective, zhu2022multi}. However, issues have been pointed out where empathy definitions are absent or abstract, and emotional empathy is overemphasized, while cognitive empathy is often absent or minimized \cite{lahnala2022critical}. 

NLP models for recognizing empathy typically treat empathy as a classification or regression task. However, this introduces a gap: in clinical settings, speaking with empathy is often viewed as \textit{aligning} the Observer's speech to the Target's, yet we lack methods for how to explicitly identify this alignment.
Our work directly addresses this gap by recognizing cognitive and emotive appraisals \cite{smith2010theory,lamm2007neural,appraisalWondra} and measuring empathy in terms of the degree of Observer alignment with a Target's situation and appraises it in the same way.

\section{A Dataset of Empathetic Appraisals}

To facilitate research on cognitive and emotional empathy, we introduce a new dataset of Target and Observer pairs, \dataset (\textbf{Al}igment \textbf{o}f \textbf{E}mpathy), annotated for how each appraised the Target's situation and which appraised passages are aligned.

\subsection{Data Source}
\label{sec:data-source}

Data was drawn from Reddit, which hosts a diverse range of communities focused on mental, emotional, and social support \citep{de2014mental,gkotsis2016language}. Support typically occurs in two settings. Most commonly, an individual in need of support with make a post describing their situation, and then others may reply in comments to the post; additionally, a user may comment in a conversation thread that solicits a supportive discussion, e.g., a weekly post requesting such comments.  Candidate data for annotation was selected from all post-comment pairs and comment-reply to those posts in 35 English-language subreddits (Appendix \ref{app:subreddits}) from 2019-01 to 2021-06. This collected 28,018  post-comment and 1367 comment-comment candidate pairs for annotation. 

Not all content in these communities relates to empathy, e.g., off-topic conversations or posts from moderators. To focus specifically on empathy-related content, we pre-filter data using the models of \citet{zhou2020condolence}; their models identify content relating to distress, whether a reply is condolence, and an ordinal measure of the empathy of a reply. Details of these classifiers are in Appendix \ref{app:data-filtering}. We retain only annotation candidates where (1) the post was classified as distress and the reply as condolence and (2) the empathy rating for the reply was $\ge$2, on a scale from $[1,5]$. This latter constraint was designed to prioritize content likely to have empathetic appraisals, as the majority of replies are low-empathy. Finally, we discard pairs where the Target contained $\ge$ 3 uses of ``you" to avoid cases where the Target was itself a response to other distress posts or comments.\cuttext{\footnote{When replying, some Observers describe their own distressing situation in detail as a way of relating to the Target, which ultimately gets classified as a distress comment.}} In total, 29,385 Target-Observer pairs were collected.

\subsection{Annotation Task and Process}
\label{sec:annotation_process}

Our annotation process consisted of extensive pilot work to develop annotation guidelines and multiple rounds of annotation and discussion. 

\myparagraph{Tasks}
Two annotation tasks were performed. The first asked annotators to highlight spans of the Target's and Observer's texts that matched one of 9 categories. Here, we include the six appraisal categories proposed by \citet{appraisalWondra}, described in Appendix~\ref{app:annotation-instructions}. Our initial pilot work identified three other categories that warranted annotation. 
Target often includes some description of the situation that is neutral with respect to their appraisal, which we label as \objexp or they may actively ask for advice from others (\advice). Observers, in turn, may also share similar experiences (\objexp), provide suggestions or advice (\advice), or use sympathetic tropes such as ``I'm sorry for your loss'' (\textit{Trope}). We include these additional span types as (1) they each reflect a common category of response type seen in everyday language---not just Reddit, (2) their inclusion helps annotators distinguish each construct from the appraisals, and (3) they offer a new way to model empathetic alignment beyond appraisals and provide more structure for understanding the lived experiences of how people receive social support, e.g., by identifying how others empathize (or struggle to) in their responses.
Examples spans of these appraisals are shown in Appendix Table \ref{tab:appraisal_example}.

Annotators were allowed to highlight spans of varied length, from clauses to multiple sentences, depending on how the individual wrote. Annotators were instructed to label a passage with only a single span type; if a sentence contained multiple span types, each should be marked separately. %

The second task had annotators align the spans between Target and Observer. Annotators were shown all labeled spans of the first phase and asked to identify any pairs where the Observer's span references a Target. An Observer span was allowed to be aligned to multiple Target spans, as often the Observer attempts to summarize and synthesize what the Target has said in their response. Full annotation instructions for both tasks are described in Appendix \ref{app:annotation-instructions}

\myparagraph{Annotation Process}
The annotation process is divided into two phases: annotating the spans of appraisals, and annotating the alignment of spans between Target and Observer. 
Due to the complexity of the task, annotators were recruited in person to receive training. Five annotators participated and went through six hours of training using the annotation codebook reported in Appendix \ref{app:annotation-instructions}. Following training, annotators worked and met weekly to discuss controversial annotations across annotators. Annotators used a custom web interface to annotate (Appendix \ref{sec:ann_web_interface_sec}), which also allowed them to take notes on specific instances they wanted to discuss, which were used to improve the codebook when applicable. Phase 1 annotations were completed in batches of 634 instances.

Phase 2 alignment annotations were completed by 4 annotators who were also involved in producing the labels for Phase 1. Annotators used a custom web interface shown in Appendix Figure \ref{fig:align_ann_interface} following a separate codebook for deciding when spans were aligned. %

\myparagraph{Adjudication Process}
In both phases, following each batch's completion, annotators participated in a review and adjudication process where all were allowed to compare their annotations with others, leave comments on why they labeled certain appraisals, and make changes to annotations of their own will. This process was designed to let annotators have access to different mindsets from others, as interpreting appraisals can be subjective based on one's own way of understanding the situation. Once Phase 1 annotation was complete, all remaining disagreements were resolved by one expert annotator prior to starting Phase 2. Following the completion of Phase 2, one expert annotator resolved all remaining disagreements on alignment. 

Because of adjudication, we do not report IAA, as this is not a meaningful estimate of reliability. Annotating appraisals is challenging due to the perspective-taking required, and adjudication was essential for mutual conceptualizing and agreeing upon the likely appraisals in many cases. We describe the challenges later in Section \ref{sec:annotation-difficulty}. 

\myparagraph{Annotated Dataset Summary}
Annotators ultimately identified 9,284 spans across 636 Target-Observer pairs, with 3,262 alignments across spans. Table \ref{tab:stats_appraisalset} shows the appraisal counts for both Target and Observer, and how many times a Target's appraisal was aligned with an Observer span.

\begin{table}[!t]
\centering
\resizebox{0.47\textwidth}{!}{
\rowcolors{2}{gray!12}{white}
\begin{tabular}{rccc}
\textbf{Span Type} & \textbf{Target}  & \textbf{Observer} & \begin{tabular}[c]{@{}c@{}}\textbf{Has alignment} \\ \textbf{in Observer}\end{tabular} \\ 
\hline
Pleasantness                               & 1059 & 487 & 522 \\
Situational Control                        & ~~744 & 278 & 268 \\
Anticipated Effort                         & ~~738 & 357 & 273 \\
Self-other Agency                          & ~~906 & 507 & 465 \\
Certainty                                  & ~~798 & 541 & 393 \\
Attentional Activity                       & ~~223 & ~~40 & ~~74 \\
Objective Experience                       & ~~885 & 362 & 168\\
Advice                                     & ~~137 & 857 & 103\\
Trope                                      & ~~~~~~0 & 363 & ~~~~0\\
\end{tabular}
}
\caption{Statistics of \dataset dataset.}
\label{tab:stats_appraisalset}
\end{table}

\subsection{Challenges in identifying appraisals}
\label{sec:annotation-difficulty}
Three common themes in difficulties were encountered during annotation, described next.

\myparagraph{Implicit Expressions} Some emotions are inferential and implicit in the text. For example, a user may say ``My cat died yesterday", which would be considered  \textit{Pleasantness} if we infer the likely emotion experienced. However, due to the distressing content, many such passages would be rated for inferred Pleasantness and so we opt to only rate explicit mentions of emotion.

\myparagraph{Ambiguity} The language of some spans was sufficiently ambiguous to elicit multiple appraisal types, e.g.``Depression in relationships can be tough." The phrase ``tough" could be interpreted with respect to emotion (\textit{Pleasantness}) or the amount of effort needed \textit{Anticipated Efforts}.

\myparagraph{Descriptions of Attention}
Among all appraisal types, \textit{Attentional Activity} was most difficult to distinguish due to the infrequency with which Targets explicitly focus on their surprise or focus of attention; instead, such language is used to indicate other types of appraisals that are more dominant in their salience, leading to its rarity in our data.

\section{Classifying Appraisals and Alignment}

Models were trained to identify spans of appraisals and to align spans between Target and Observer.

\subsection{Appraisal Prediction}
We first performed the task of automatically annotating appraisals in both Target and Observer. Due to its rarity, we excluded the \textit{Attentional Activity} type from our model and set them to be \textit{No Label} in this task. Most of the annotated spans were whole sentences\camerareadytext{, except the case where sub-sentences showed observable different appraisals,} so we predicted at the sentence level\camerareadytext{, i.e. given a Target or Observer text containing $l$ sentences: $\langle s_1, s_2, \ldots, s_l\rangle$, each $s_i, 1\le i\le l$ is passed to the model independently to be predicted.} When multiple appraisals were present, we selected the longer span in terms of characters and, when equal in length, arbitrarily broke ties. 
We combined data from Target and Observer when training models.

Classification models were trained starting from pre-trained language models (PLMs): BERT-large-uncased \cite{devlin2019bert}, RoBERTa-large \cite{liu2019roberta}, SpanBERT-large-cased \cite{joshi2020spanbert}, DeBERTa-v3-cased \cite{he2023debertav3}, sentence-transformers/all-MiniLM-L6-v2 \cite{wang2020minilm}. We also tested using a prompt-based models: OpenPrompt+BERT-large-uncased \cite{ding2021openprompt, devlin2019bert}, OpenPrompt+RoBERTa-large \cite{ding2021openprompt, liu2019roberta}, OpenPrompt+T5-large \cite{ding2021openprompt, 2020t5}. Additional training details are reported in Appendix \ref{app:model-details}.
The baseline was set as the random prediction.

\myparagraph{Results}
In general, prompt-based models performed better than PLMs, as shown in Table \ref{tab:appraisal_model_perf_simple}, and all models outperformed the baseline. Examining appraisal-level performance (Appendix Table \ref{tab:align_model_perf_big}), we saw that \textit{Advice},  \textit{Trope}, and \textit{Objective Experience} were the easiest to classify, while \textit{Anticipated Effort} as the lowest. However, classification performance was similar for most appraisal types, indicating the model was sufficiently effective to label data for large-scale analysis.

\cuttext{Compared to the baseline, all models exhibited the ability to capture the appraisals. The results also reflected the complexity of language that drives ambiguity. A typical example is shown below whose prediction was from OpenPrompt+RoBERTa.
\begin{quote}
    \textbf{Text:} There are some things you just cannot unsee.\\
    \textbf{Label:} Situational Control \\
    \textbf{Model Prediction}: Certainty
\end{quote}
``Cannot unsee'' can be interpreted as someone having no control over choosing to see "some things" or not (Situational Control), or can be interpreted as someone must see "some things" based on the double negative.}

\begin{table}[!]
\resizebox{\columnwidth}{!}{
\rowcolors{2}{gray!12}{white}
\begin{tabular}{rccc}

& \textbf{F1} & \textbf{Recall} & \textbf{Precision} \\
\hline

\textit{random}                                                         & 0.11                                                & 0.11                                                    & 0.11                                                       \\
\textit{majority} & 0.03 & 0.02 & 0.11 \\
BERT                                                           & 0.38                                                & 0.38                                                    & 0.41                                                       \\
RoBERTa                                                        & \textbf{0.56}                                                & 0.56                                                    & 0.57                                                       \\
SpanBERT                                                       & 0.52                                                & 0.52                                                    & 0.54                                                       \\
DeBERTa                                                        & 0.55                                                & 0.55                                                    & 0.56                                                       \\
MiniLM                                                         & 0.49                                                & 0.50                                                    & 0.51                                                       \\
OpenPrompt+BERT     & 0.53                                                & 0.53                                                    & 0.55                                                       \\
OpenPrompt+ RoBERTa  & \textbf{0.56}                                                & \textbf{0.57}                                                    & 0.58                                                       \\
OpenPrompt+ T5-large& \textbf{0.56}                                                & 0.56                                                    & \textbf{0.59}                                                       \\
\end{tabular}
}
\caption{Appraisal model performance.}
\label{tab:appraisal_model_perf_simple}
\end{table}

\subsection{Alignment Prediction}
Alignment prediction between Target and Observer was done using a Siamese Network \cite{siamese_network}. 
The task is structured as, given a span of text from the Target and a span of text from the Observer, predict whether the Observer's appraisal is aligned with the Target's appraisal.
\camerareadytext{Formally, given a pair of Target and Observer $(T, O)$ with annotated appraisals/spans where $T=\{\text{span}_{t_1}, ..., \text{span}_{t_k}\}$, $O=\{\text{span}_{o_1}, ..., \text{span}_{o_l}\}$, the input data is $T\times O$ with label $Y\in \{0, 1\}^{k\times l}$.}
Because most pairs did not align, \camerareadytext{and the alignment between three pairs (\advice and \objexp, \advice and Pleasantness, Anticipated Effort and \objexp) does not exist or is extremely rare (fewer than 7 occurrences)}, when constructing the dataset, we omit those pairs and downsampled to a positive-negative ratio of 1:11.
We tested using the all-MiniLM-L6-v2 or all-mpnet-base-v2 parameters to initialize the Siamese Network. The baseline was set as picking a random label with the empirical distribution of the dataset. We evaluated the performance using Binary F1 for is-aligned.

In addition to these two trained models, we also include two other baselines that focus just on text similarity: threshold classifiers trained on either the Jaccard Index of the words in the two passages or on a Siamese network with all-mpnet-base-v2 parameters. Both baselines allow us to test whether empathetic alignment is simply textual similarity, or, as theory predicts, a deeper alignment that goes beyond content.

\myparagraph{Results}
Both Siamese network models were able to meaningfully identify alignment, as shown in \tref{tab:align_model_perf}, with the mpnet-base-v2 model performing best. 
Notably, both threshold-base baselines show that empathetic alignment requires more than text overlap or semantic similarity between two spans; while both baselines do attain high precision, they fail to recognize the majority of the cases where the Observer is aligning.
However, alignment classification is a challenging task, with our best model only attaining a binary F1 of 0.46. %
In particular, the task requires significant social reasoning capabilities to understand how an Observer's speech is reflective of the Target's description, which provides significant room for improvement. Appendix table \ref{tab:align_wrong_examples} shows examples highlighting the variety and subtly in determining alignment.

\begin{table}[t]
\centering
\resizebox{0.48\textwidth}{!}{
\rowcolors{2}{gray!12}{white}
\begin{tabular}{rcccc} 
\textbf       &  \textbf{Recall} & \textbf{Precision} & \textbf{F1} \\ \hline
\textit{random}              &  0.09          & 0.08           & 0.08        \\
baseline: word overlap       &  0.01          & 0.55           & 0.02 \\
baseline: all-mpnet-base-v2  &  0.02          & 0.62           & 0.04 \\
fine-tuned all-MiniLM-L6-v2  &  0.41          & 0.42           & 0.41 \\
fine-tuned all-mpnet-base-v2 &  \textbf{0.45} & \textbf{0.46}  & \textbf{0.46}        \\
\end{tabular}
}
\caption{Alignment model performance.}
\label{tab:align_model_perf}
\end{table}

\subsection{Appraisal and Alignment Dataset}

We applied our best model (OpenPrompt+RoBERTa) to predict appraisals in comments and posts in 91 subreddits relating to mental health and support (listed in Appendix \ref{sec:big_reddit_tree_subreddits}). For each post, we classified whether the post was about distress using the approach described in Section~\ref{sec:data-source} and then labeled the appraisals for the post and all comments made under that post.
After combining the consecutive sentences that were predicted to have the same appraisal, we passed them to all-mpnet-base-v2  for alignment prediction. %
We applied this pipeline of models to 2.3M posts and 8.9M comments, identifying 21.7M appraisals in Targets' posts or comments and 326.9M appraisals in Observers' comments.
We used this dataset for all analyses.

\section{Appraisal Behavior}
\label{sec:appraisal-behavior}
Different types of distressing events may be more likely to evoke specific appraisals, such as (un)pleasantness for the loss of a loved one, or the effort involved to handle mental illness. 
The 91 communities in our data cover a range of possible situations and \citet{littleBlackBox} note that empathy must be understood in context, with responses that adapt to the circumstances.
Here, we test whether individuals in these communities show regularity in how they appraise as Targets and, do Observers, in turn, vary the appraisals with which they respond.

\myparagraph{Setup}
PCA is then run on a matrix of subreddits and their normalized distribution of appraisals across all their posts.

\begin{figure}[!t]
    \centering
    \includegraphics[width=0.47\textwidth]{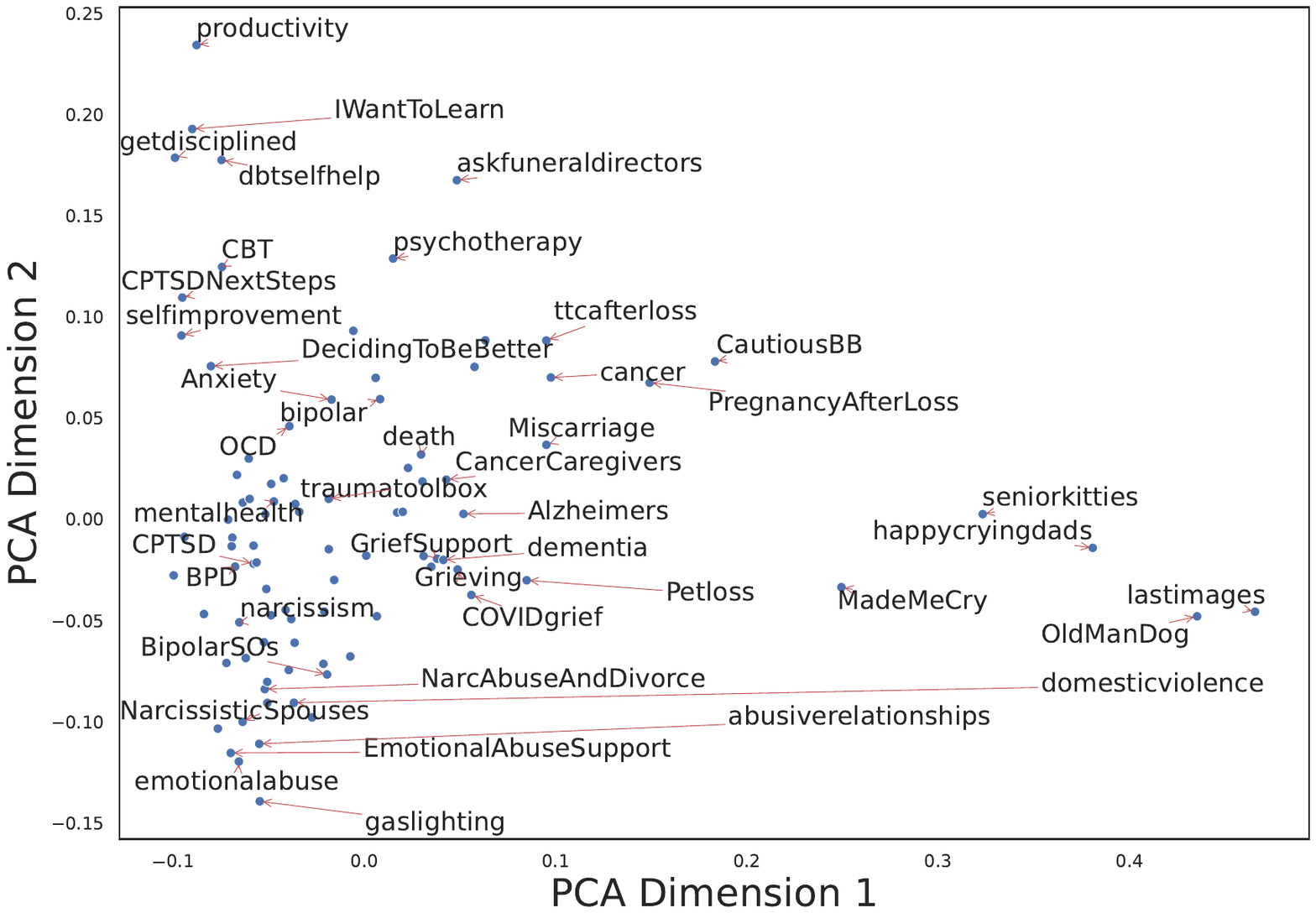}
    \caption{Subreddits arranged according to their distribution of Target appraisals. }
    \label{fig:target-pca}
\end{figure}

\begin{figure}[!t]
    \centering
    \includegraphics[width=0.47\textwidth]{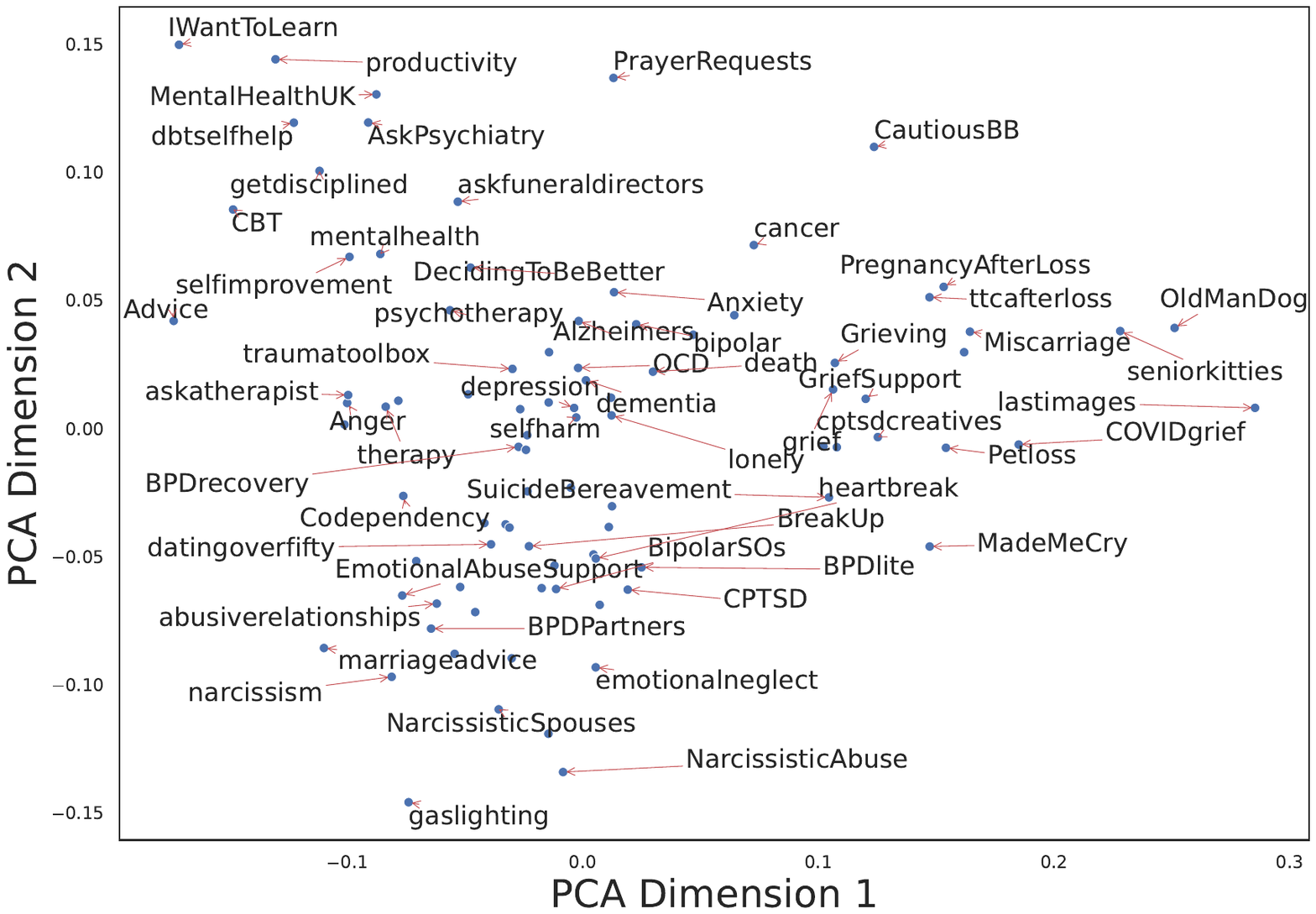}
    \caption{Subreddits arranged according to their distribution of Observer appraisals. }
    \label{fig:observer-pca}
\end{figure}

\myparagraph{Results}
Communities were thematically clustered solely based on the relative distribution of appraisals (not content), shown in Figure~\ref{fig:target-pca} and \fref{fig:observer-pca}. 
For example, for Targets, clusters are seen for communities focused on structured therapy and self-help modalities (top left), recipients of abuse (bottom left), and various topics of grief (center to center right). Similar clusters are also seen based on how Observers appraise in these subreddits.
Although these subreddits all contain distressing situations, there is no a priori reason to expect that they should differ in how people categorize their lived experiences---many distressing situations could easily be described with any of the appraisals, yet individuals show regularity in how they rationalize and describe their thematically-similar experiences.
This emergent grouping suggests that the related situations that targets find themselves in between these communities, while different, lead to similar ways of appraising those situations\cuttext{---even though any situation may be described with any type of appraisal}. 
The behavioral differences in how Observers appraise from our large-scale observational results support the lab study of \citet{stellar2020profiles} who found that Observers vary the themes of their responses based on the type of suffering described by the Target.

\section{Do Observers Align?}

Targets experience responses as highly empathetic when an observer appraises the situation in the same way \citep{vyskocilova2011empathy,watson2016role}, which requires that their appraisals align with those of the Target. Given the behavioral similarities seen between Targets and Observers in which appraisals they use, here we test whether the responses align.

\myparagraph{Setup}
We calculate the probability that an Observer $O$'s appraisal of type $a_j$ is aligned the each type $a_j$ when used by the Target $T$: $p(a_i^O|a_j^T)$.

\begin{figure}[!t]
    \centering
    \includegraphics[width=0.47\textwidth]{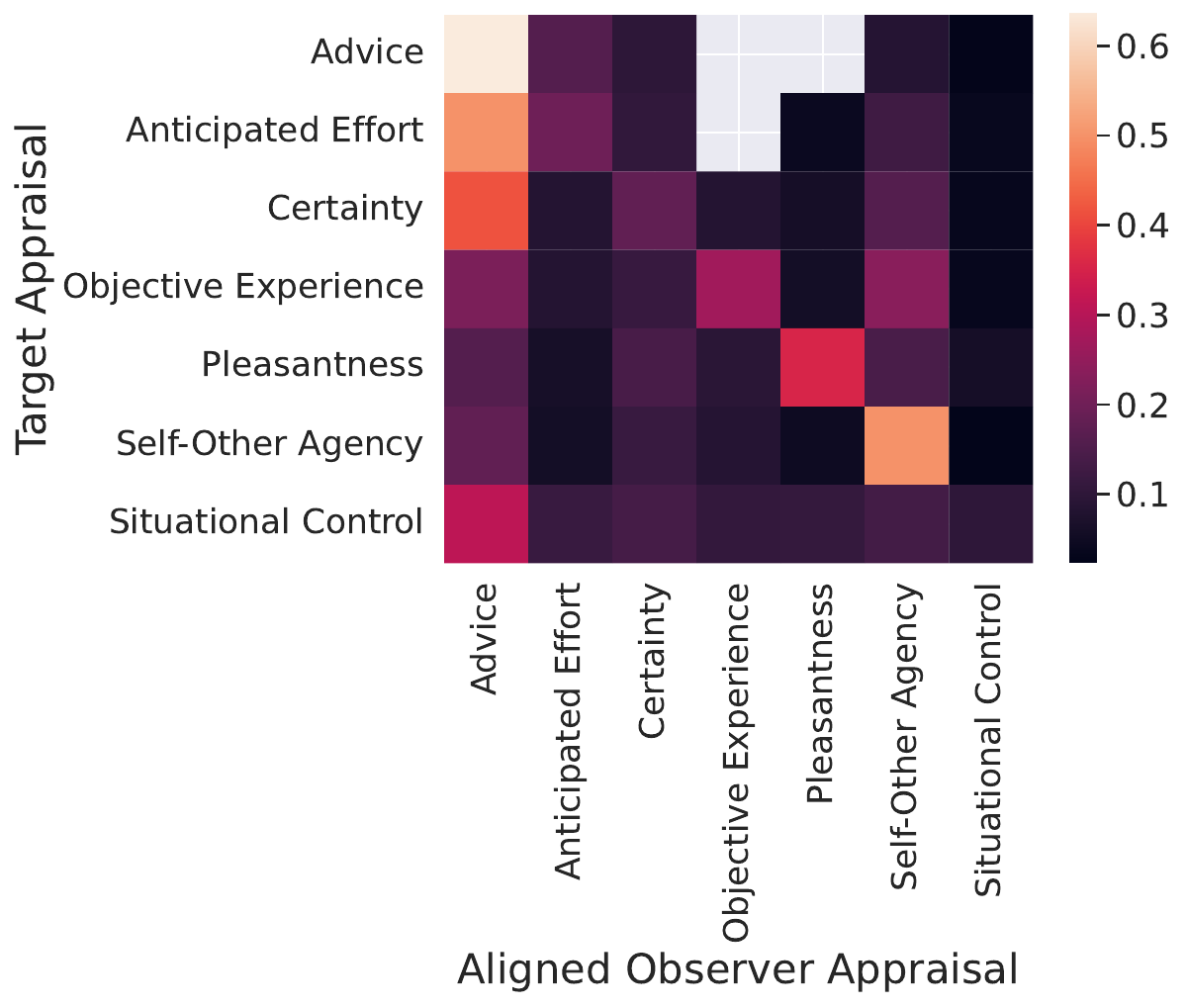}
    \caption{The probabilities of a Target's appraisal type (row) having the specific appraisal (col) in the aligned span of the Observer's. Empty cells indicate the aligned pairs are rare or non-existent in the data. }
    \label{fig:alignment}
\end{figure}

\myparagraph{Results}
In aggregate, Observers only partially aligned with how the Targets appraised (experienced) their situation (\fref{fig:alignment}). Instead of having the same appraisal, the majority of the Observer's aligned text was giving advice to the Target about a particular aspect. 

Giving advice is a well-known aspect of Reddit support communities \citep[e.g.,][]{de2014mental} and some individuals so seek out communities for such advice \citep[e.g.,][]{sowles2017feel,o2018today}. While advice is not considered a component of empathy---and in some circumstances is considered counter-productive when used in empathetic situations like counseling \citep{barkham1986counselor,lieberman1999bathe}---its frequency does highlight its importance in the lived experience of support-seeking individuals. Indeed, \citet{depow2021experience} note that the experience of empathy in everyday situations encompasses a much broader set of behaviors than those listed in academic definitions.

Nevertheless, we do see a strong diagonal trend in \fref{fig:alignment} that suggests that, when not giving advice, Observers do frequently align in their appraisals, suggesting empathetic behavior. Two off-diagonal trends also emerge. First, Observers frequently respond with Certainty; in our annotation, we found that these were frequently gestures meant to reassure the Target of their choice or action, and, thus, may be viewed as a type of compassionate response. Second, Observers often respond to a Target describing the objective experience with comments about the Target's agency (or not) in the situation; here too we find a type of compassion-based response where Observers use agency language to deflect responsibility from the Target onto other parties mentioned in the Target's experience.
Appendix \ref{app:subreddit-analysis} reports additional details on specific subreddits' differences and behaviors.

\section{Alignment by Professional Observers}

Subreddit communities contain a mixture of Observers, some of whom have professional training in mental health or medical domains. Such training frequently includes discussions of empathetic and patient-centered dialog \citep{hojat2016empathy,lam2011empathy}. Some communities allow users to include a flair next to their username to indicate a self-reported qualification, such as a PhD in Psychiatry. Given that users with such flairs should have experienced some training on how to behave more empathically---i.e., more alignment---here, we test the level of alignment of different professions of Observers relative to the general public in our data.

\myparagraph{Setup} 
We used the flair on Reddit as an indicator of whether a user is a professional or not. We adopt the flair-profession categorization of \citet{lahnala2021exploring} to map the text of each flair to a specific profession:\footnote{We note that some professions overlap in their theme. For example, Psychiatrists, Psychologists, Psychotherapists, and Social Workers may all be considered Therapists or Counselors. However, some qualifications, such as a Licensed Professional Counselor (LPC) do have an associated degree. } Counselor, Funeral Role, Medical Doctor, Nurse, Psychiatrist, Psychologist, Psychotherapist, Social Worker, or Therapist; see Appendix \ref{app:flairs} for details. Flairs not indicating a specific degree or known license were left unmapped. For those with professional degrees, we also extract any reported status in training as either Fully Licensed or a Student. 
Professionals are defined as those who are licensed and have a non-student title, while laypeople are authors who do not have any student or professional flair throughout their usage of Reddit. We further filtered the replies from the professionals when they were of their highest/most recent training level. Ultimately, we identified 14,648 users as professionals, 2978 as students, and 1,686,362 laypeople in our data for analysis. Appendix Table \ref{tab:profession-counts} shows the count by profession.

Alignment is measured as the percentage of appraisals in the Target's message that have an alignment in the Observer's. We report the mean percentage for laypeople and each profession's users, calculated over all data in our dataset.

\begin{figure}[!t]
    \centering
    \includegraphics[width=0.47\textwidth]{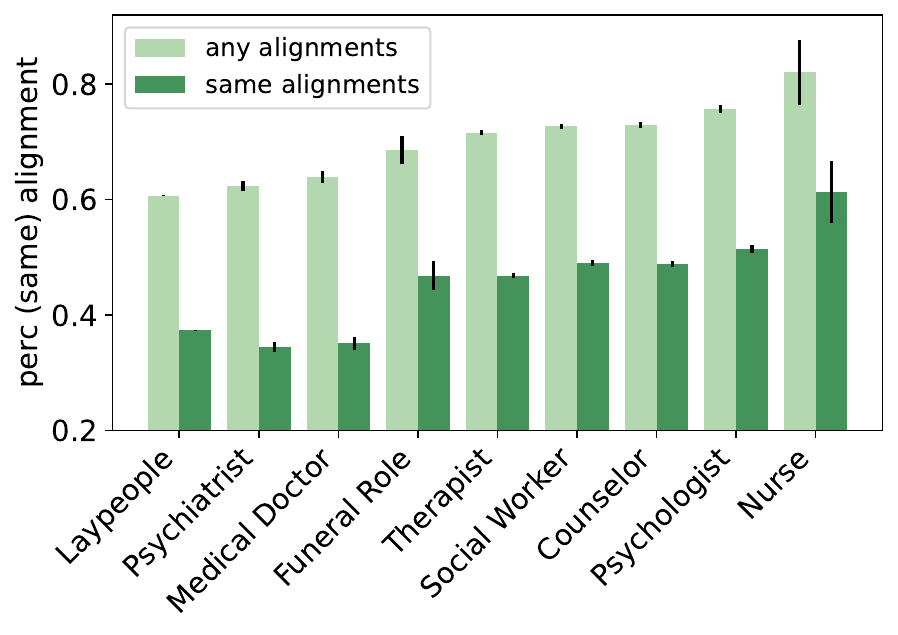}
    \caption{Mean alignment by profession; the error bar for laypeople is too small to be seen.}
    \label{fig:profession-alignment}
\end{figure}

\myparagraph{Results} 
Mental health professionals have higher alignment than laypeople, shown in \fref{fig:profession-alignment}, both for aligning with the same appraisals and in general. A small split can be seen within professions as well: Professions for clinical therapy (Therapist, Social Worker, Counselor, and Psychologist) have among the highest alignment with Targets, while medical Professions (Psychiatrists and Doctors) are much lower---even lower than laypeople at matching the same alignment. Our results suggest that the training received by mental health professionals does lead to higher alignment than laypeople. %

\paragraph{Does the flair itself drive behavior?}

Individuals who list their professional degrees as flair in a subreddit often interact with others in different subreddits where no such flair is visible. With no explicit mention of their profession, there is less reputational harm in responding with less effort which could lead to lower alignment.  To test whether flair visibility drives alignment, we fit a linear regression on the Observer's percent of alignment with categorical factors for (i) the profession, (ii) the subreddit, and (iii) profession flair visibility.

Flair visibility leads to a small but significant increase in alignment ($\beta$=0.027; p<0.01); full regression results are in Appendix Table \ref{tab:is_title_visible}. However, the magnitude of the increase suggests that professionals still reply with relatively high empathy when not publicly sharing their profession. \camerareadytext{Note that this effect is also not due to the change in community, as the subreddit regression factor controls for relative differences in alignment between communities.}

\paragraph{How do professionals differ in alignment?}

Given that mental health professionals better align with Targets, we examine how they differ from laypeople in which appraisals are aligned. Here, we restrict professionals to the most aligned: Therapist, Social Worker, Nurse, Psychologist, and Counselor as professionals in this analysis, while laypeople were defined as before. 
To control for potential differences in the content Observers are responding to, we only examine Target messages that have replies by at least one professional and one layperson. Replies were grouped by professionals and laypeople, with the percentage of the same appraisal alignment over each Target appraisal calculated. 
To compare professionals and laypeople, we calculate the difference in mean probability of using the same appraisals as the Target.

\begin{figure}[!t]
    \centering
    \includegraphics[width=0.47\textwidth]{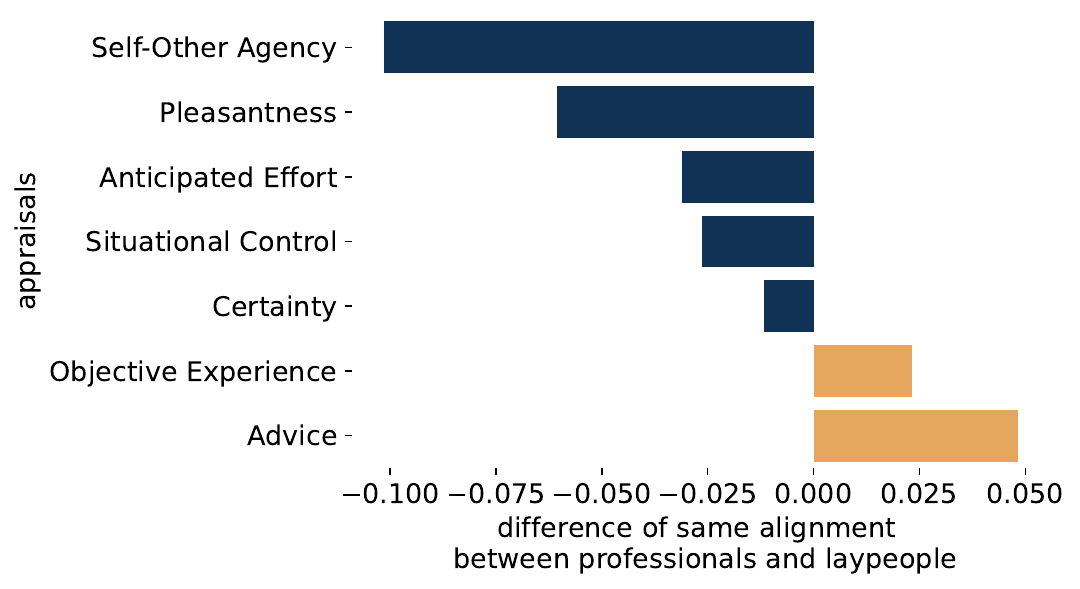}
    \caption{Difference of mean alignment between professionals and laypeople by appraisal; bars pointing right are appraisals more commonly aligned by professionals, left for laypeople.}
    \label{fig:difference_prof_laypeople_same_conv}
\end{figure}

Surprisingly, while professionals have higher total alignment, they are much less likely to use the same appraisals in their response (\fref{fig:difference_prof_laypeople_same_conv}). Controlling for Target, professionals are much less likely to respond to Target's appraisals about their agency or the situation's pleasantness with the same appraisal, compared with laypeople. Instead, we find their aligned responses are more commonly advice to the Target.

\section{Does experience influence alignment?}

Professional health practitioner training emphasizes the importance of empathetic communication. However, multiple studies have noted that this training period marks a high point, and doctors and nurses become less empathetic over time \cite{wilson2012empathy,chen2007cross}. Our first question is whether we observe a similar drop-off after therapist students transition to their fully licensed roles in our longitudinal data.

Laypersons too may benefit from explicit feedback on what comments are considered helpful, likely learning how to engage more empathetically over time. Redditors are known to offer such feedback when the response matches their support-seeking goals \cite{peng2021effects}. As a second related question, we test whether laypersons become more empathetic as they receive such feedback on which comments were most helpful.

\myparagraph{Setup}
For the first question, we collect all authors who have flairs with licensed or student training levels. We then collected all of their engaged conversations as observers and split their comments into licensed and student periods based on when the flair text changes. Comparison is made between conversations involving licensed observers and student observers. 
For the second question, we use data from \textit{r/Advice}, which assigns flairs of different levels to users based on how many times they have replied as an Observer and another user has replied to express gratitude for their comment. We treat these flairs as proxies for experience in writing helpful replies. We collected the replies from different experience levels (flairs) and calculated the mean percentage of alignment for each level. \cuttext{The results are shown sorted from low experience to high experience.}

\begin{figure}[!t]
    \centering
    \includegraphics[width=0.47\textwidth]{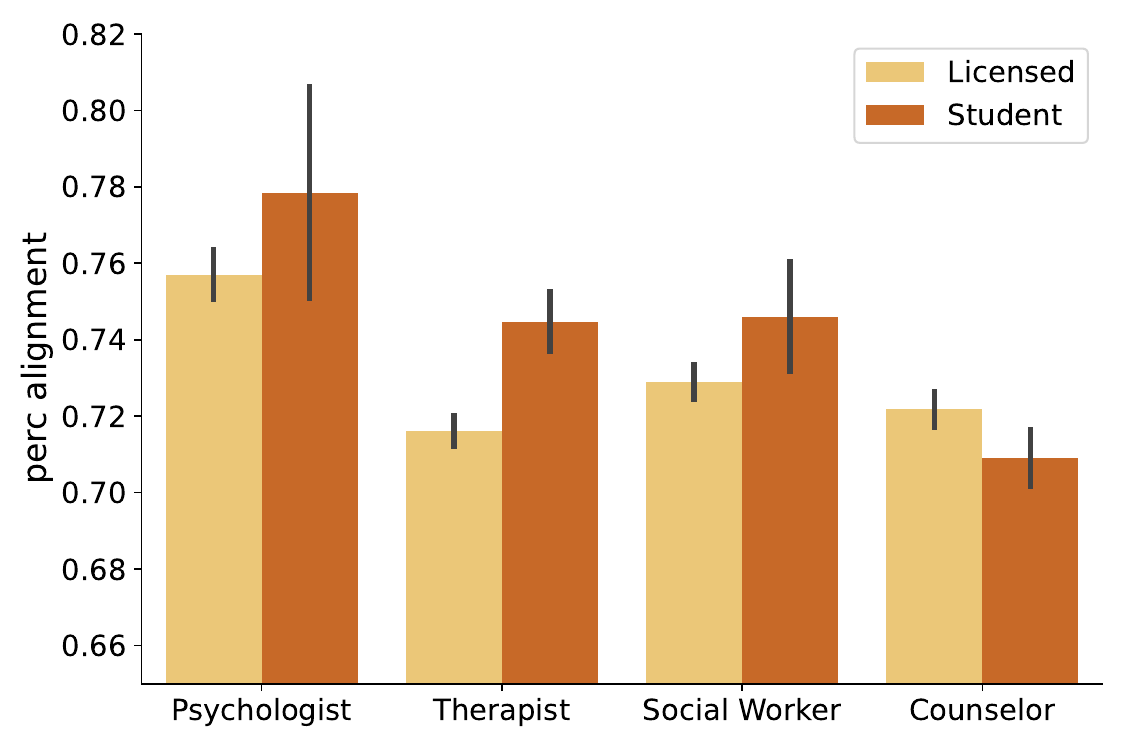}
    \caption{Mean alignment comparison between the licensed and students}
    \label{fig:license_student}
\end{figure}

\myparagraph{Results} 
Our results show that students are often more empathetic than their fully licensed counterparts (\fref{fig:license_student}). Of these, only the drop for Therapist users is significant at p$<$0.1 using an independent t-test. Our observations mirror results from \citet{wilson2012empathy} and \citet{chen2007cross} showing that nursing and medical students decreased their empathy levels with patients as they received more training. One likely driver of such drop-off is compassion fatigue, where high levels of empathy towards Targets in a therapeutic setting can lead to a decreased ability to feel compassion for others \citep{turgoose2017predictors}.

\begin{figure}[!t]
    \centering
    \includegraphics[width=0.47\textwidth]{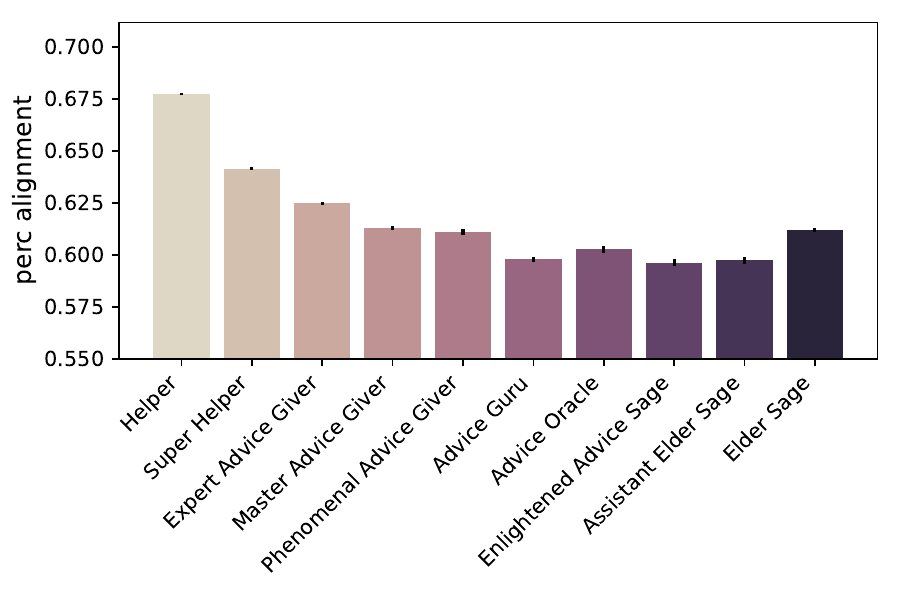}
    \caption{Mean alignment by the amount of gratitude received at the time of comment, shown as flair from r/Advice ordered from least to most thanked. }
    \label{fig:advice_level}
\end{figure}

A similar trend is seen for users in r/Advice as they gain more experience making helpful comments (\fref{fig:advice_level}), where users initially comment with very high alignment and then slowly drop in alignment during their continued engagement until they reach a state-state level.
While status-seeking is a known strong motivator for Reddit users \citep{moore2017redditors}, 
we hypothesize that as users engage more frequently, some drop off in empathetic alignment may come from social media fatigue, which can decrease psychological well-being \cite{dhir2018online} and thus lead to a decreased ability to empathize. We also hypothesize, that similar to professionals,  these users also experience compassion fatigue from engaging with distressing comments.

\section{Discussion}

Individuals seek support in online communities, yet the type of support they receive varies, and, as we show, may not be well-aligned or, when aligned, may be advice rather than validation of their experience. The result that healthcare professionals are more likely to give advice raises new research questions about how online and offline therapeutic practices may differ. Future studies could examine (i) the qualitative differences in the advice of laypersons vs professionals, (ii) whether Targets view this advice as more valuable or empathetic, and (iii) if possible, the underlying motivations for professionals to give more advice, despite their training. The models developed in this paper can help surface such examples for study.

The task of empathetic alignment can provide new opportunities for advancing NLP.
For example, generating empathetic responses is effortful for many people  \cite{cameron2019empathy}, in part, because of the mental work. NLP models for recognizing this alignment could be used for assistive technologies that lower the cognitive load for responding with high empathy, such as highlighting passages that an Observer might respond to and assessing whether their responses match what a Target has actually said. Such tools could potentially provide lower-effort entry points into the conversation to help people engage.

\section{Conclusion}

Empathy requires perspective taking on the part of an Observer to align their cognitive and emotional experiences with another. This study goes beyond prior work in NLP on empathy to make these empathetic alignments explicit and to identify how observers mirror (or miss) the types of perspectives described by Targets.  By developing a new dataset, \dataset, and models for appraisals and alignments in empathetic dialogues, our work enables studies of how and when Observers empathize. In a large-scale study of Reddit, we show that individuals seeking mental health support do receive empathetic replies---but that many aligned responses are giving advice, rather than acknowledging their perspective. However, we also show that mental health professionals on Reddit show much higher alignment than the general public. Our data and model can support future studies on how to help identify and correct misalignment when drafting responses or suggest opportunities for new alignments.
All data, models, and annotation materials are available at {\small{\url{https://github.com/jessicayjm/modeling_empathy_alignment}}} and the annotation tool is available in a stand-alone form at {\small{\url{https://github.com/jessicayjm/span_alignment_annotation_tool}}}

\section{Limitations}

Our study examines conversations on a public social media platform, Reddit. Thus, our results may not be generalizable to other settings such as in-person settings where modalities other than texts might also be significant, or where individuals can speak in full confidence of anonymity. However, having the benefits of using longer texts, we were able to perform analysis on more complex appraisals compared with methods using shorter texts such as Tweet data or text messages.

Secondly, we only analyzed Reddit posts and the top-level comments and replies to those posts. While these post-comment pairs offer the cleanest signal of individuals looking for mental health support, our focus necessarily limits empathetic conversation that may be happening in replies to comments. We view this as an opportunity for future work in multi-turn dialog in Reddit. Meanwhile, analysis from the Reddit data still carries its significance for longer conversation analysis given the impressive number of users being active on the platform.

A key limitation in annotation comes from the inaccessibility to the mindsets of original targets and observers. Our dataset reflects third-party perceptions of the state of mind of annotators---a challenging task given that we lack information on who the targets and observers are. However, under the design of the annotation process, our annotations likely mirrored the process that the observers go through when assessing potential targets. Further, our adjudication process ensured that multiple potential interpretations were considered when any message was ambiguous so that the most likely could be chosen. Future work should be encouraged to collect first-hand annotations directly from targets and observers, as this is currently a missing dataset for the community.

We should also be aware of the inherent biases of lived experience from annotators. Though crowdsourcing could provide more diversity, the annotation task itself is complex and not immediately amenable to crowdwork without extensive validation. We hope that our work can set a baseline for further generalization and to test how annotator identity can influence perceptions of empathy and alignment.

The general trends in our experiments rely on classifiers that imperfectly learn how to identify appraisals and how Targets and Observers align. Given the challenge of these classification tasks, our initial models attain only moderate performance which could potentially influence our downstream results.
As a result, we have taken care to only describe trends in aggregate and to report confidence intervals and standard errors wherever possible. While moderate in performance, our approach mirrors work in other NLP tasks such as framing \citep[e.g.,][]{ajjour2019modeling,akyurek2020multi,van2020context,dayanik2022improving,mendelsohn2021modeling} where models operate on nuanced, often social, data to identify a moderate number of labels in order to derive large-scale trends when applying these classifiers at scale. Nonetheless, social information remains challenging to recognize even for large language models \cite{choi2023llms} and our dataset provides an opportunity for future work to improve performance at recognizing empathetic alignment, which can open new doors for more fine-grained analyses of empathetic behavior.

Last, our results are drawn from a primarily Western social media context and a Western-educated annotation pool. This cultural backdrop likely limits the generalizability of our results to other cultures. In our work, the annotators were aware of the cultural context of the paper's data. Other work will be needed to understand how individuals from a variety of cultural backgrounds appraise distressing settings and how they effectively engage with empathy.

\section{Ethical Considerations}

The work includes some comments by people who have experienced or are experiencing distressing events. While this data is fully public---posted by the authors themselves publicly to seek support---additional views of the data could risk having them being further re-exposed to the events by malicious actors. However, we view this risk as very low compared with the potential benefits of studying distressing events by providing insights for helping Observers better engage with Targets, which would lead to long-term support in the community. 

Considering the data source is public but sensitive, we only release the data to researchers after filling out a request that acknowledges the potentially-sensitive nature of the data and the responsibility of its use.

\section*{Acknowledgments}

The authors thank reviewers for their thoughtful and valuable feedback on the paper. We thank Allie Lahnala for her feedback and sharing the Reddit flair resources. We also are \textit{extremely} grateful for the wonderful annotators who worked with us on this project: Allyson Elwart, Lauren Kim, Wendy Liu, Hayley Cho, and Jacky He.  This work was supported by the National Science Foundation under Grant Nos. IIS-2007251 abd IIS-2143529.

\bibliography{anthology,custom}
\bibliographystyle{acl_natbib}

\appendix

\section{Subreddits Used for Data}
\label{app:subreddits}

\subsection{Subreddits for building alignment dataset} \textit{anxiety, depression, Miscarriage, domesticviolence, widowers, GriefSupport, Petloss, FiftyFifty, SuicideBereavement, ttcafterloss, heartbreak, BreakUps, BreakUp, BipolarSOs, dementia, Alzheimers, ExNoContact, CautiousBB, domesticviolence, CaregiverSupport, abusiverelationships, emotionalabuse, marriageadvice, lastimages, PrayerRequests, OldManDog, seniorkitties, askfuneraldirectors, death, dogpictures, MadeMeCry, cancer, MomForAMinute, sad, happycryingdads}

\subsection{Subreddits for Reddit tree}
\label{sec:big_reddit_tree_subreddits}
\textit{depression, BPD, dementia, Vent, abusiverelationships, offmychest, lonely, BreakUps, SOCIALSKILLS, CautiousBB, Advice, TalkTherapy, MomForAMinute, adultsurvivors, getdisciplined, MadeMeCry, NarcissisticAbuse, bipolar, SuicideWatch, Anxiety, widowers, selfharm, GriefSupport, BPDlovedones, SingleParents, Anger, mentalhealth, datingoverforty, heartbreak, emotionalabuse, ExNoContact, lastimages, PrayerRequests, PregnancyAfterLoss, marriageadvice, DecidingToBeBetter, SuicideBereavement, CPTSD, socialanxiety, seniorkitties, IWantToLearn, OldManDog, Petloss, ttcafterloss, cancer, psychotherapy, OCD, datingoverfifty, emotionalneglect, Alzheimers, BorderlinePDisorder, Codependency, self-improvement, death, gaslighting, BPDPartners, productivity, dbtselfhelp, CaregiverSupport, NarcAbuseAndDivorce, MMFB, therapy, Miscarriage, domesticviolence, BipolarSOs, BreakUp, askatherapist, sad, LifeAfterNarcissism, AskPsychiatry, FriendsOver40, NarcissisticSpouses, ChildrenofDeadParents, BPDlite, happycryingdads, CBT, narcissism, Grieving, BodyAcceptance, MentalHealthUK, BPD4BPD, askfuneraldirectors, InternalFamilySystems, CPTSDNextSteps, EmotionalAbuseSupport, CancerCaregivers, cptsdcreativeas, BPDrecovery, grief, traumatoolbox, COVIDgrief}

\section{Pre-Annotated Data Filtering}
\label{app:data-filtering}
We used all three models (Distress, Condolence, and Empathy classifiers) from \citet{zhou2020condolence} to filter the Reddit data. Both distress and condolence classifiers are bert-base-uncased models, while the empathy classifier is a roberta-base model. Filtering was performed to surface data likely to contain empathy.
We first identify all posts where p(distress)$>$0.9, then we retain comment replies rated as p(condolence)$>$0.9. From these post-comment pairs, we retain all pairs with an empathy rating of at least 2 on their 5-point scale. 

\section{Additional Annotation Details}

\subsection{Annotation Instructions}
\label{app:annotation-instructions}

Annotators were instructed to read the 11-page annotation codebook (included in the Supplementary Data). The codebook contains detailed instructions and examples for each appraisal type and the three new span types we introduce. Table \ref{tab:appraisal-defs} shows the general definition in the codebook for each.

\begin{table*}[t]
    \centering
    \rowcolors{2}{gray!12}{white}
    \begin{tabular}{lp{0.7\textwidth}}
        \textbf{Appraisal/Span} & \textbf{General Definition} \\ \hline
        Pleasantness & How pleasant the situation was. \\
        Anticipated Effort & How much effort was needed to deal with the situation. \\
        Situational Control & How much the situation was out of anyone’s control. \\
        Self-other Agency & How much oneself or another person was responsible for the situation. \\
        Attentional Activity & Reflects how much the Target’s attention was drawn to the situation rather than diverted away from the situation. This appraisal has more to do with the occurrence of an event---its suddenness, familiarity, and predictability---rather than the qualities of the event like its pleasantness. \\
        Certainty & Certainty about what was happening in the situation or what would happen next \\
        Objective Experience &  Description of the experience of the author that is not an appraisal or the broader context/circumstances in which their story takes place. \\
        Advice & Expressions of asking or providing advice. \\
        Trope & General sympathetic expressions that are not specific to the Target. \\

    \end{tabular}
    \caption{General definitions used in the annotation codebook for each of the appraisal types. In the codebook, each definition is followed by more details, notes, and examples of what is or is not that appraisal for both observers and targets.}
    \label{tab:appraisal-defs}
\end{table*}

\subsection{Annotator Recruitment}

Annotators were recruited from a large mailing list of university undergraduates.  Interested undergraduates participated in an initial paid one-hour training session and five (4 women, 1 man) signed on to continue annotating after an initial vetting of their work done during the training session. Annotators were paid \$15/hr USD and were able to work up to 10 hours per week, with flexibility depending on their schedule.

\begin{table}[!h]
\centering
\rowcolors{2}{gray!12}{white}
\begin{tabular}{rrrr} 
 & Train & Dev  & Test \\ \hline
No Label                   & 461   & 49   & 133  \\
Pleasantness                    & 956   & 124  & 229  \\
Anticipated Effort                   & 767   & 113  & 198  \\
Certainty                    & 934   & 163  & 281  \\
Objective Experience                   & 1146  & 144  & 357  \\
Self-Other Agency                   & 1161  & 252  & 369  \\
Situational Control                   & 782   & 112  & 229  \\
Advice                    & 1120  & 181  & 321  \\
Trope                    & 262   & ~~31   & ~~80   \\ \hline
\textit{Total}                & 7589  & 1169 & 2197 \\
\textit{Multi-label}          & ~~872   & ~~130  & ~~234  \\ 
\end{tabular}
\caption{Number of sentences for appraisal prediction containing both Target and Observer. Multi-label shows how many sentences contain more than one appraisal.}
\label{tab:align_data_stats}
\end{table}

\subsection{Annotation Website Interface}
\label{sec:ann_web_interface_sec}

\begin{center}
\begin{figure*}[!h]
    \centering
    \includegraphics[width=\textwidth]{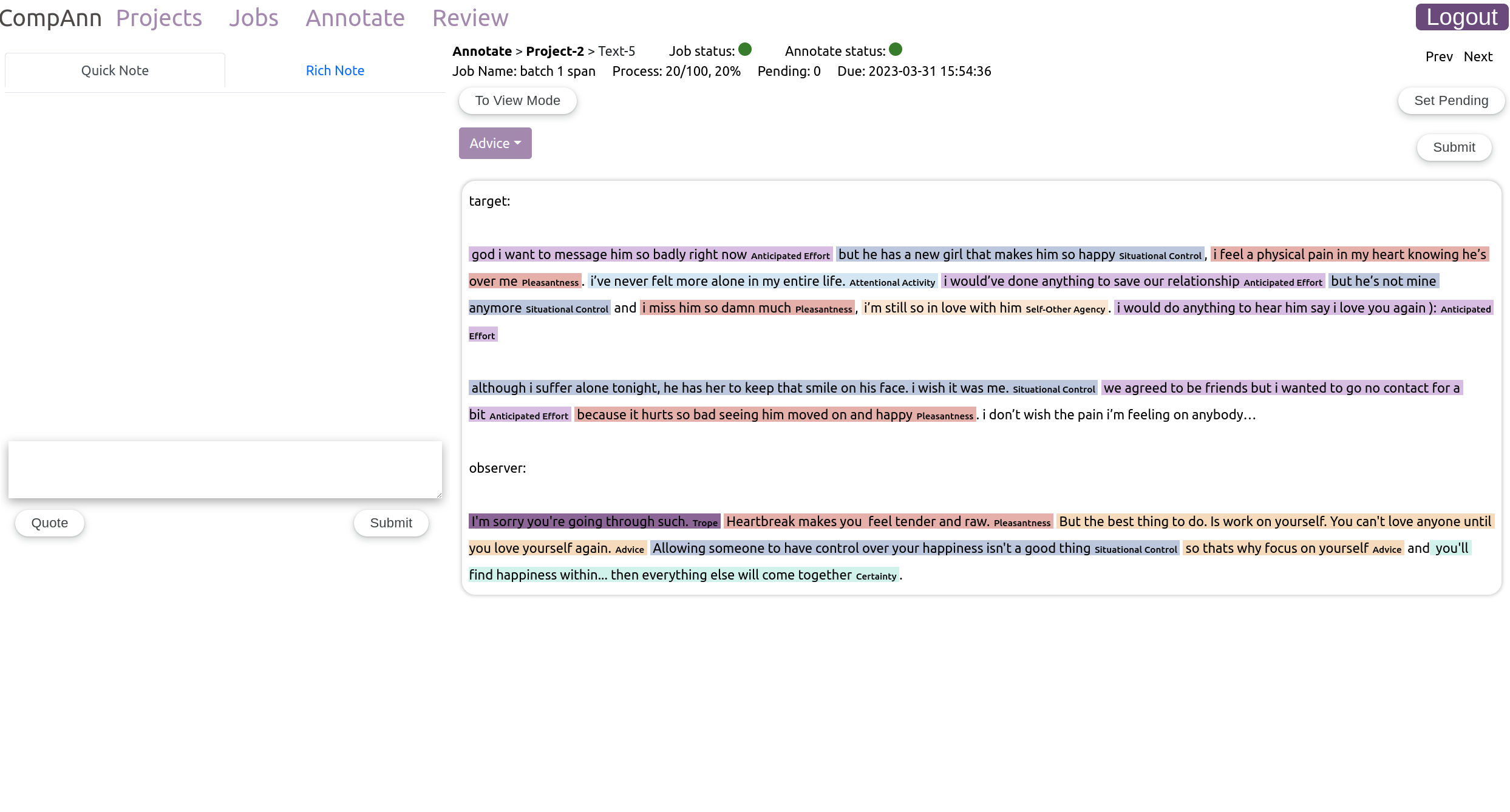}
    \caption{Appraisal Annotation interface.}
    \label{fig:span_ann_interface}
\end{figure*}
\end{center}

\begin{center}
\begin{figure*}[!h]
    \centering
    \includegraphics[width=\textwidth]{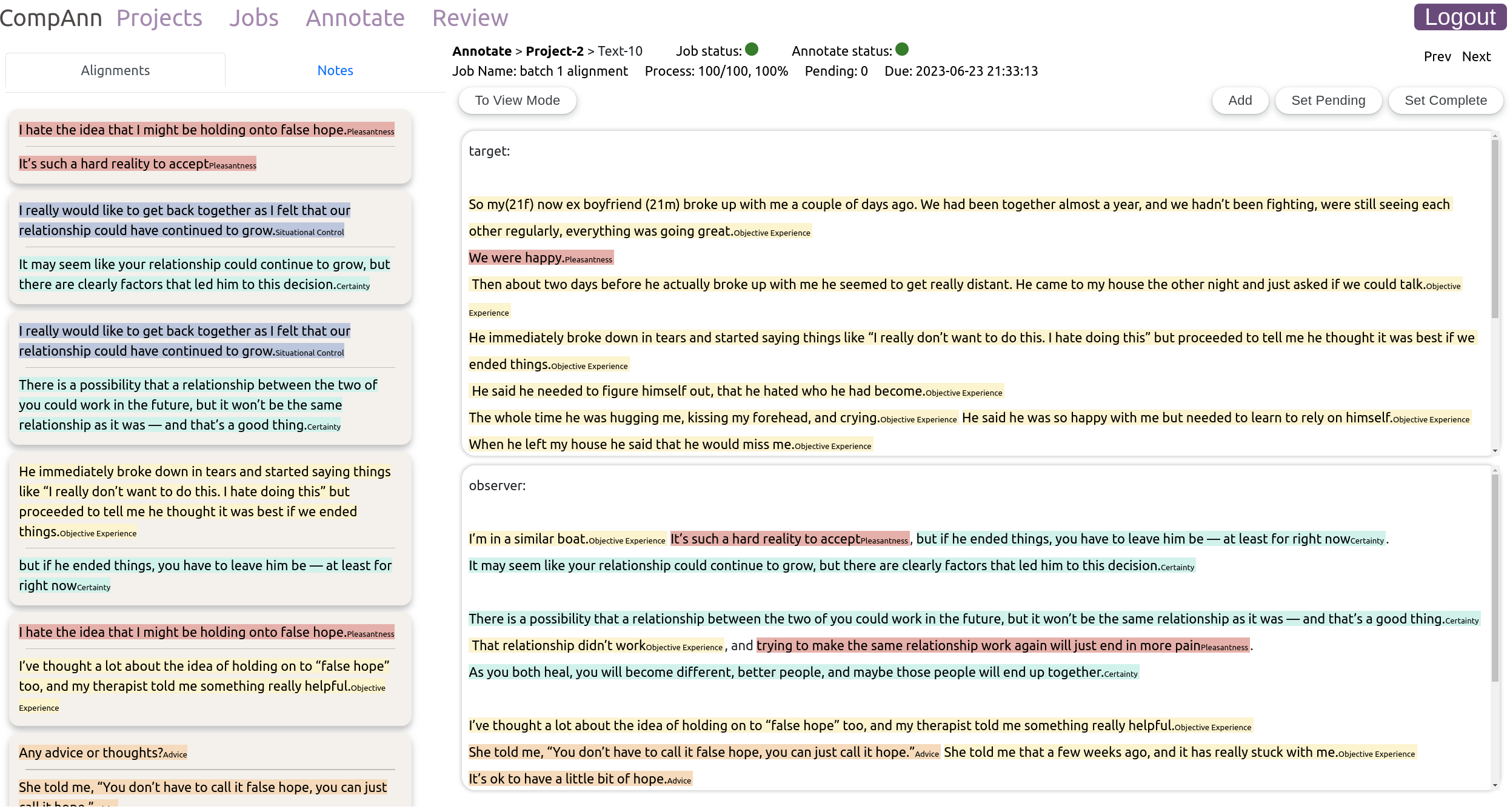}
    \caption{Alignment Annotation interface.}
    \label{fig:align_ann_interface}
\end{figure*}
\end{center}

\begin{center}
\begin{figure*}[!h]
    \centering
    \includegraphics[width=\textwidth]{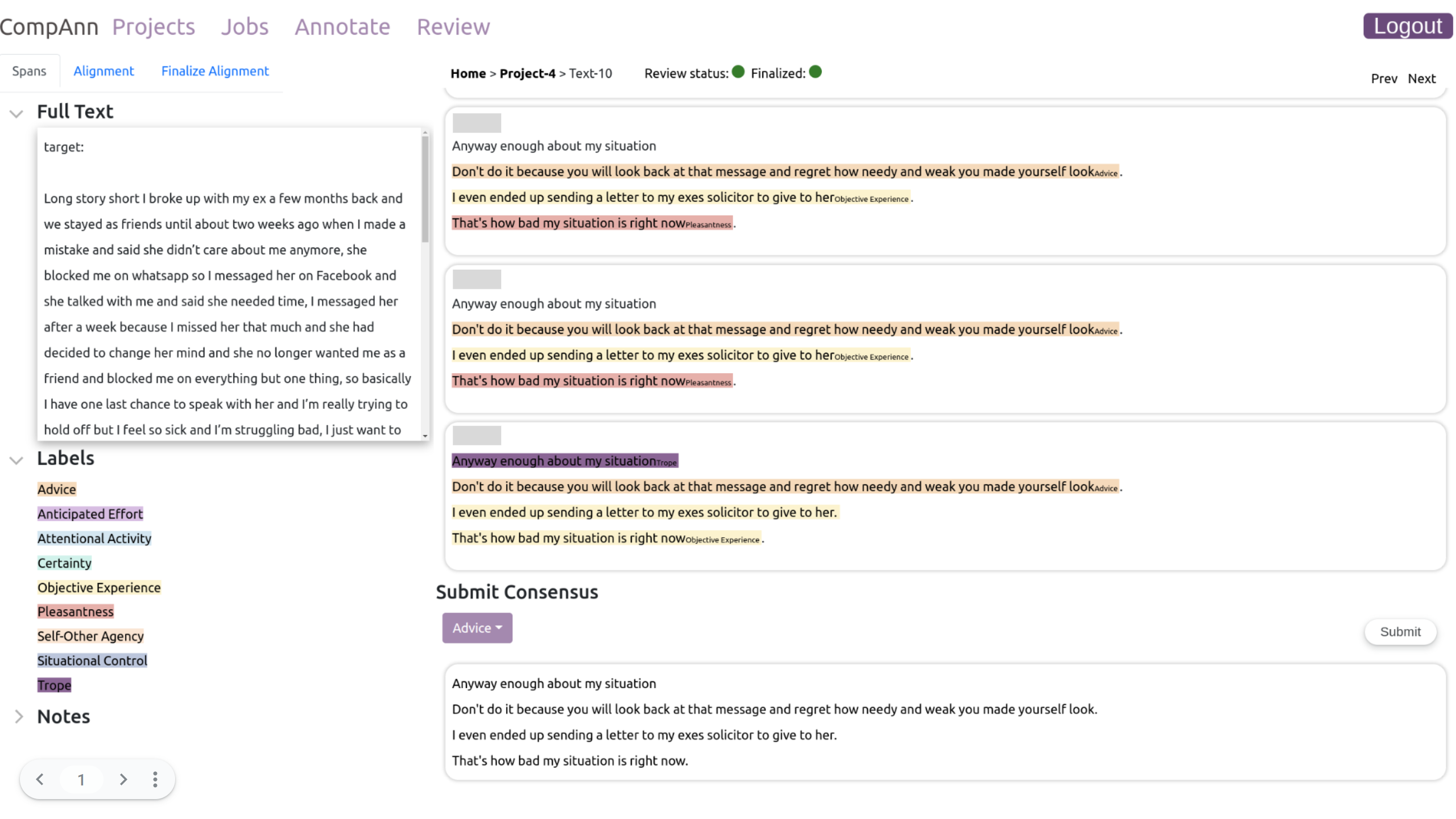}
    \caption{Finalize appraisal annotation interface.}
    \label{fig:finalize_span_ann_interface}
\end{figure*}
\end{center}

\begin{center}
\begin{figure*}[!h]
    \centering
    \includegraphics[width=\textwidth]{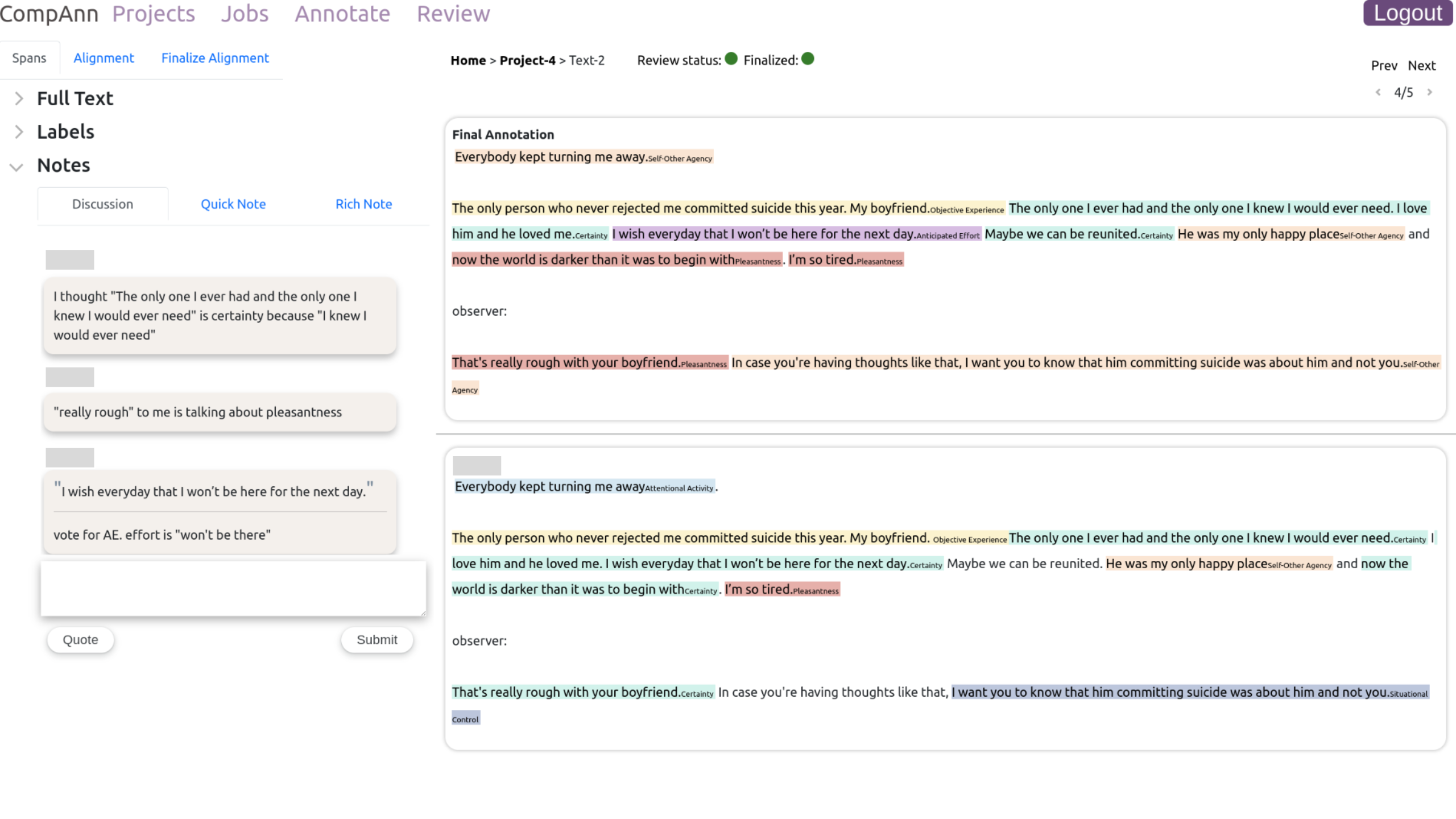}
    \caption{Finalize alignment annotation interface with note function.}
    \label{fig:finalize_span_ann_interface_w_note}
\end{figure*}
\end{center}

\begin{center}
\begin{figure*}[!h]
    \centering
    \includegraphics[width=\textwidth]{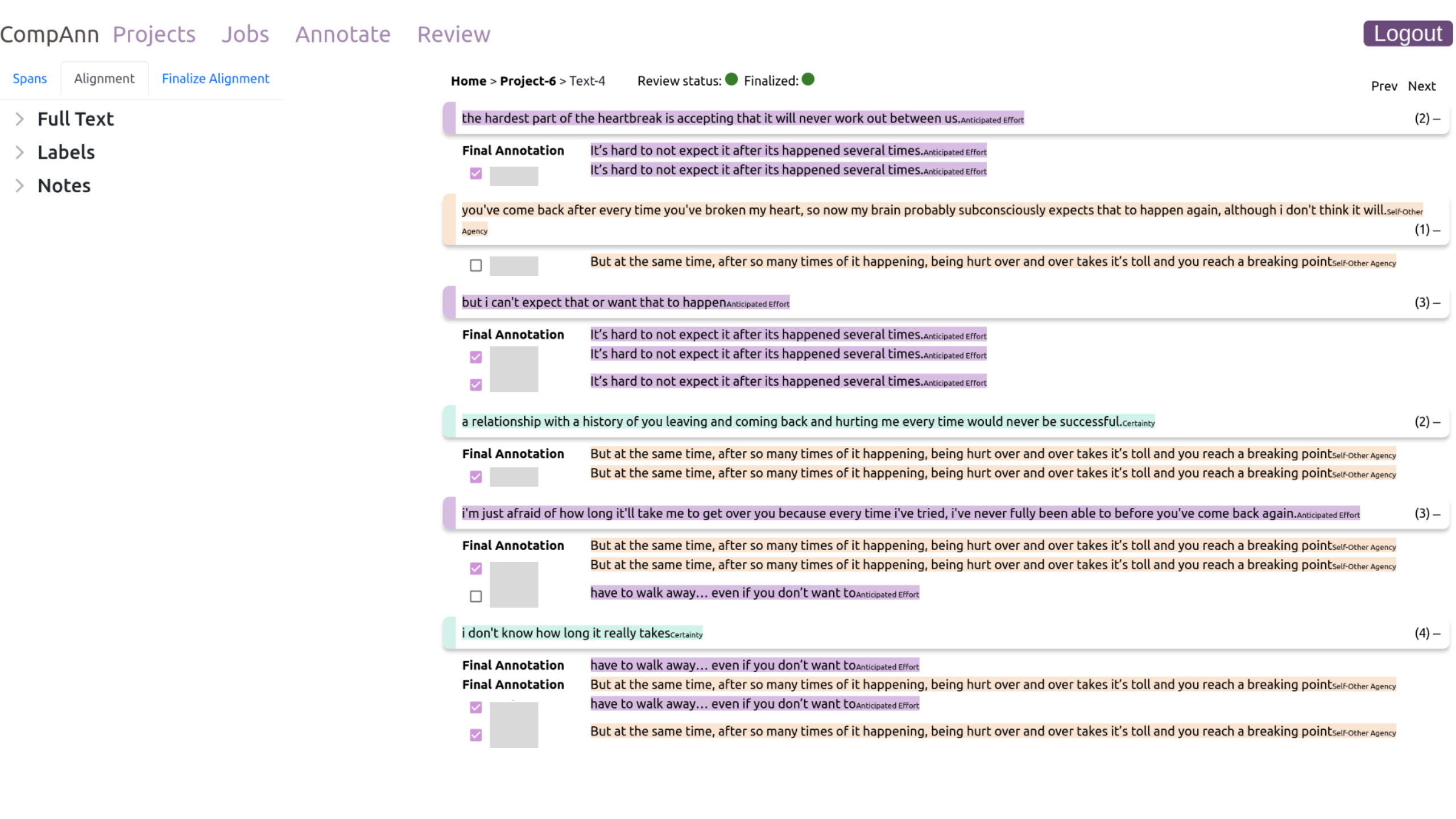}
    \caption{Finalize alignment annotation interface: view and select from annotators.}
    \label{fig:view_all_alignment_annotations}
\end{figure*}
\end{center}

Annotators used a custom website for all their work. The interface for appraisal annotation is shown in \fref{fig:span_ann_interface}. Annotators can select labels and highlight spans for annotation, track the annotation process, and make private notes.

The interface for alignment annotation is shown in \fref{fig:align_ann_interface}. Annotators can click one span from Target and one span from Observer to annotate alignments. The interface shares the note function with the appraisal annotation interface.

The review interface for both appraisal and alignment is the same as the annotation interface but with an extra discussion function where annotators can post their comments, raise questions regarding each instance, and communicate with each other asynchronously.

The interfaces for the admin user to finalize the appraisal annotations are shown in \fref{fig:finalize_span_ann_interface} and \fref{fig:finalize_span_ann_interface_w_note}. The admin has access to all annotators' work and is able to decide the final annotation. The discussion panel shows all public comments from annotators.

The interfaces for the admin user to finalize the alignment annotations are shown in \fref{fig:view_all_alignment_annotations}. The admin can directly select alignment from the annotators' work. New alignments that are not identified by annotators can be added through \textit{Finalize Alignment} panel which is the same as the alignment annotation interface.

\subsection{Additional Annotated Examples}

Table \ref{tab:appraisal_example} shows additional examples of appraisals and other categories that were annotated.

\subsection{Addition Observations on Labeling}

Attentional Activity was rare in our data, in part, because the general perception was that other types of appraisals were more salient and likely explanations. For example, a strongly \textit{Attentional Activity} dominated span could be: "On the one hand I don’t want to go around starting every conversation announcing that my brother has passed, but it’s been THE central event in my life recently and the biggest thing on my mind.'' However, in many cases, other appraisals will dominate the interpretation, such as in the following examples: 
\begin{itemize}
    \item \textit{Pleasantness} dominates: ``I’ve never felt more alone in my entire life."
    \item \textit{Anticipated Effort} dominates: ``Depression was and is still the hardest challenge that I face everyday."
    \item \textit{Objective Experience} dominates: ``Called her this morning and police picked up saying she is dead."
\end{itemize}

\begin{table*}[t!]
\begin{center}
\rowcolors{2}{blue!12}{white}
\begin{tabular}{p{\textwidth}}
\hline \textbf{Appraisals} \\ \hline\hline
(\textit{Pleasantness}) I feel so absolutely shattered into infinite pieces and even that doesn't seem to express this enough. \\
(\textit{Anticipated Effort}) I know it’s important to live in the now but right now I’d rather sleep than face the reality of anything. \\
(\textit{Situational Control}) \\ Whenever I let myself feel joy in life again my brain says "hey imagine if you were feeling this joy with your ex, remember that?" and my heart starts to hurt again. \\
(\textit{Self-Other Agency}) She broke up with me, she destroyed my heart and she's the one finding someone new to be happy with while i haven't had even a crush on anyone in all this time. \\
(\textit{Attentional Activity}) I never imagined it would be this hard to cope, especially for a pregnancy that ended so early. \\
(\textit{Certainty}) I don’t know if I ever wanted the relationship to begin with or if I just wanted validation. \\ 
\end{tabular}
\vspace{2mm}
\rowcolors{2}{white}{red!12}
\begin{tabular}{p{\textwidth}}
\hline \textbf{Non-appraisals} \\ \hline\hline
(\textit{Objective Experience}) I posted about my sweet girl Lulu just a bit over a month ago. Since then her thyroid tumor masses have gotten so big that they have started compressing her throat and now she can’t eat. \\
(\textit{Trope}) I’m so sorry you have to go through this. \\
(\textit{Advice}) Sometimes it helps to share your good memories, either in places like this one, or with other family/friends who knew him. \\
\hline
\end{tabular}
\end{center}
\caption{Examples on appraisals and non-appraisals. }
\label{tab:appraisal_example}
\end{table*}

\section{Additional Model Details and Results}

\subsection{Training Details}

Information on the size of the different dataset splits is shown in \tref{tab:align_data_stats}, as well as what percent of the data was originally labeled with multiple spans of different appraisal types (11\%). 

All parameters not mentioned use default values in Huggingface transformers library \cite{wolf2020huggingfaces}. The random seed is set to be 0 for all the training.

\subsection{Span prediction}
All span prediction models are trained on cross-entropy loss with AdamW optimizer \cite{loshchilov2019decoupled}. OpenPrompt+RoBERTa-large is trained with a learning rate of 1e-7, while all other models are trained with a learning rate of 1e-6. 
Specifically for OpenPrompt models: \texttt{freeze\_lm=False}, \texttt{max\_seq\_len=512}, \texttt{decoder\_max\_len=3}, \texttt{teacher\_forcing=False}, \texttt{truncation\_method=head}.
All other specific information on the training process and hyperparameters are shown in Table \ref{tab:train_span_info}.

\begin{table*}[!tb]
\centering
\rowcolors{2}{gray!12}{white}
\begin{tabular}{rccccc}

& \textbf{max epoch} & \textbf{patience} & \textbf{batch size} & \textbf{best epoch} & \textbf{time}\\
\hline
BERT  & 200 & 15 & 32 & 23 & 02:35\\
RoBERTa  & 200 & 15 & 32 & 10 & 01:39 \\
SpanBERT   & 200 & 10 & 32 & 14 & 01:38 \\
DeBERTa  & 200 & 10 & 16 & ~~7 & 01:47 \\
MiniLM &  200 & 10 & 16 & 38 & 00:22 \\
OpenPrompt+BERT  & 200 & 10 & 16 & 11 & 03:56 \\
OpenPrompt+ RoBERTa &  200 & 20 & 16 & 52 & 14:02  \\
OpenPrompt+ T5-large &  200 & 10 & ~~8 &57 & 21:37 \\
\end{tabular}
\caption{Training information for predicting appraisals.}
\label{tab:train_span_info}
\end{table*}

\subsection{Alignment prediction}

\begin{figure}[!tb]
    \centering
    \includegraphics[scale=0.13]{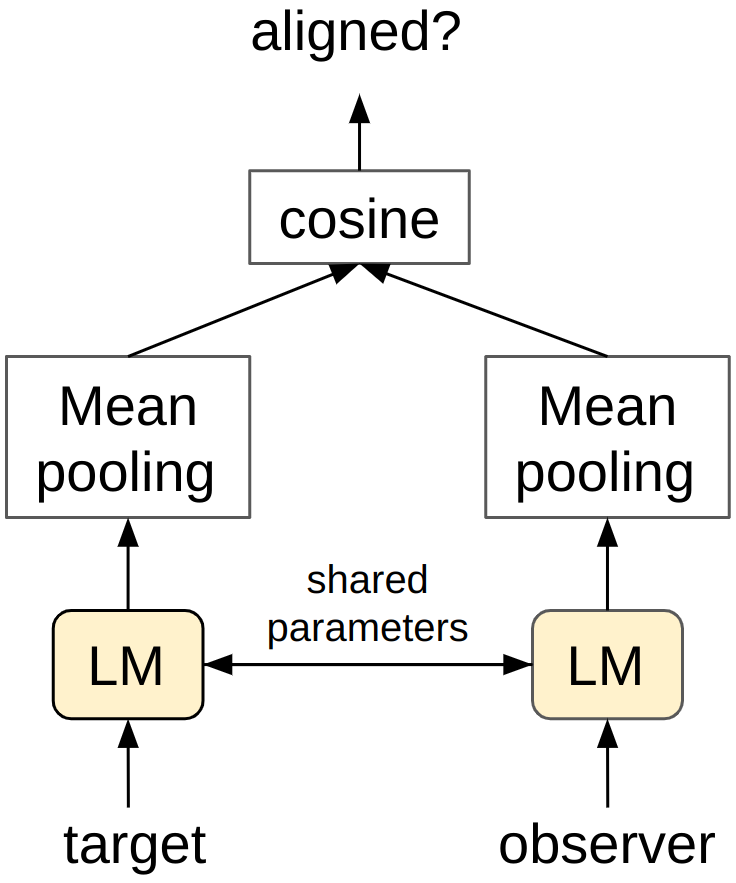}
    \caption{Alignment model architecture.}
    \label{fig:alignment_model_arch}
\end{figure}

The model architecture is shown in Figure \ref{fig:alignment_model_arch}. Both all-MiniLM-L6-v2 and all-mpnet-base-v2 are trained on mean squared error loss (mean reduction) with AdamW optimizer, \texttt{max\_epoch=300}, \texttt{patience=15}, \texttt{batch\_size=16}. The prediction threshold is set to be 0.3 ($p\ge 0.3$ will be predicted as aligned). all-MiniLM-L6-v2 reached the lowest dev loss at epoch 4 with a total training time of 1 hour and 15 minutes. all-mpnet-base-v2 reached the lowest dev loss at epoch 9 with a total training time of 5 hours and 7 minutes.

\subsection{Model Performance}

Detailed model performances for appraisal prediction are shown in Table \ref{tab:align_model_perf}.

\label{app:model-details}

\begin{table*}[!tb]
\resizebox{\textwidth}{!}{
\rowcolors{2}{gray!12}{white}
\begin{tabular}{rccccccccccc}
\hline
\multicolumn{1}{l}{} & random       & majority      & BERT                 & RoBERTa              & SpanBERT             & DeBERTa              & MiniLM               & \begin{tabular}[c]{@{}c@{}}Open-\\ Prompt\\ +BERT\end{tabular} & \begin{tabular}[c]{@{}c@{}}Open-\\ Prompt\\ +RoBERTa\end{tabular} & \begin{tabular}[c]{@{}c@{}}Open-\\ Prompt+\\ T5-large\end{tabular}  \\ \hline\hline
\textbf{Macro-F1}    & 0.11     & 0.03        & 0.38                 & \textbf{0.56}        & 0.52                 & 0.55                 & 0.49                 & 0.53                                                           & \textbf{0.56}                                                    & \textbf{0.56}                                                         \\
\textbf{Macro-Recall} &0.11&0.02 &0.38&0.56&0.52&0.55&0.50&0.53&\textbf{0.57}&0.56\\
\textbf{Macro-Precision} &0.11& 0.11&0.41&0.57&0.54&0.56&0.51&0.55&0.58&\textbf{0.59}
\\ \hline
\textbf{Per Label Recall}      & \multicolumn{1}{l}{} & \multicolumn{1}{l}{} & \multicolumn{1}{l}{} & \multicolumn{1}{l}{} & \multicolumn{1}{l}{} & \multicolumn{1}{l}{} & \multicolumn{1}{l}{} & \multicolumn{1}{l}{}                                           & \multicolumn{1}{l}{}                                              & \multicolumn{1}{l}{}                                                                  \\ \hline
No Label                   & 0.06 &   0.00              & 0.14                 & \textbf{0.34}        & 0.23                 & \textbf{0.34}                 & 0.18                 & 0.32                                                           & \textbf{0.34}                                                     & 0.25                                                                                                                \\
Pleasantness                    & 0.11    &    0.00         & 0.51                 & 0.66                 & \textbf{0.69}        & 0.66                 & 0.66                 & 0.68                                                           & \textbf{0.69}                                                     & \textbf{0.69}                                                                                                    \\
Anticipated Effort                   & 0.08   & 0.00              & 0.32                 & 0.44                 & 0.43                 & 0.41                 & 0.41                 & 0.42                                                           & 0.46                                                              & 0.48                                                                                                      \\
Certainty                    & 0.12    & 0.00             & 0.36                 & 0.48                 & 0.54                 & 0.44                 & 0.50                 & 0.46                                                           & \textbf{0.58}                                                     & 0.55                                                                                                                   \\
Objective Experience                   & 0.16      & 0.00           & 0.45                 & 0.67                 & 0.58                 & 0.55                 & 0.61                 & \textbf{0.68}                                                  & 0.58                                                              & 0.60                                                                                                                \\
Self-Other Agency                   & 0.15      & 0.18           & 0.42                 & 0.62                 & 0.62                 & \textbf{0.63}        & 0.59                 & 0.61                                                           & 0.62                                                              & 0.60                                                                                                               \\
Situational Control                   & 0.10     & 0.00            & 0.11                 & 0.31                 & 0.21                 & \textbf{0.45}        & 0.14                 & 0.20                                                           & 0.31                                                              & 0.38                                                                                                                \\
Advice                    & 0.15    & 0.00             & 0.58                 & \textbf{0.73}        & 0.69                 & 0.70                 & 0.70                 & 0.71                                                           & 0.72                                                              & \textbf{0.73}                                                                                                       \\
Trope                    & 0.04      & 0.00           & 0.56                 & 0.79                 & 0.71                 & 0.76                 & 0.70                 & 0.70                                                           & \textbf{0.80}                                                     & 0.79                                                                                                                \\ \hline
\textbf{Per Label Precision}   & \multicolumn{1}{l}{} & \multicolumn{1}{l}{} & \multicolumn{1}{l}{} & \multicolumn{1}{l}{} & \multicolumn{1}{l}{} & \multicolumn{1}{l}{} & \multicolumn{1}{l}{} & \multicolumn{1}{l}{}                                           & \multicolumn{1}{l}{}                                              & \multicolumn{1}{l}{}                                                                                \\ \hline
No Label                   & 0.11    & 0.00             & 0.38                 & 0.63                 & 0.54                 & 0.46                 & 0.44                 & 0.55                                                           & 0.64                                                              & \textbf{0.70}                                                                                                         \\
Pleasantness                    & 0.13    & 0.00             & 0.35                 & 0.56                 & 0.54                 & \textbf{0.57}        & 0.47                 & 0.51                                                           & 0.54                                                              & 0.52                                                                                                                   \\
Anticipated Effort                   & 0.10   & 0.00              & 0.25                 & 0.45                 & 0.39                 & \textbf{0.47}        & 0.40                 & \textbf{0.47}                                                           & 0.46                                                              & 0.45                                                                                                                 \\
Certainty                    & 0.11      & 0.00           & 0.37                 & \textbf{0.55}        & 0.49                 & \textbf{0.55}        & 0.46                 & 0.53                                                           & 0.47                                                              & 0.54                                                                                                            \\
Objective Experience                   & 0.11   & 0.00              & 0.52                 & 0.63                 & 0.68                 & \textbf{0.70}        & 0.66                 & 0.59                                                           & 0.69                                                              & 0.66                                                                                                                \\
Self-Other Agency                   & 0.11     & 1.00            & 0.38                 & 0.53                 & 0.51                 & 0.53                 & 0.50                 & 0.50                                                           & 0.55                                                              & 0.54                                                                                                           \\
Situational Control                   & 0.10    & 0.00             & 0.42                 & 0.51                 & 0.39                 & 0.34                 & 0.39                 & 0.48                                                           & \textbf{0.55}                                                     & 0.52                                                                                                                  \\
Advice                    & 0.11     & 0.00            & 0.44                 & 0.63                 & 0.62                 & \textbf{0.67}        & 0.63                 & 0.61                                                           & 0.66                                                              & 0.64                                                                                                                   \\
Trope                    & 0.14       & 0.00          & 0.58                 & 0.70                  & 0.68        & 0.70                 & 0.63                 & 0.71                                                           & 0.67                                                              & 0.68                                                       \\ \hline                                                   
\end{tabular}
}
\caption{Appraisal model performance.}
\label{tab:align_model_perf_big}
\end{table*}

\section{Additional Profession Results}
\label{app:flairs}

The full breakdown of user and comment counts for those users with profession flairs is shown in \tref{tab:profession-counts}. 

\begin{table}[tb]
    \centering
    \begin{tabular}{lrr}
    Profession & \#Users & \# Comments \\
    \hline
Nurse          &    3 &   23 \\
Funeral Role   &   21 &  152 \\
Medical Doctor &   24 &  712 \\
Psychiatrist   &   17 & 1221 \\
Psychologist   &  114 & 1769 \\
Counselor      &  241 & 3374 \\
Social Worker  &  338 & 4049 \\
Therapist      &  377 & 4937 \\ 
    \end{tabular}
    \caption{Counts of how many users had valid flairs associated with each profession and the number of comments associated with each.}
    \label{tab:profession-counts}
\end{table}

\tref{tab:is_title_visible} shows the full linear regression model fit for predicting the level of alignment based on a user's profession, the subreddit, and whether their profession's flair was visible at the time of comment.

\begin{table*}[!htbp] 
\centering 
\resizebox{\textwidth}{!}{
\begin{tabular}{@{\extracolsep{5pt}}lc} 
\\[-1.8ex]\hline 
\hline \\[-1.8ex] 
 & \multicolumn{1}{c}{\textit{Dependent variable:}} \\ 
\cline{2-2} 
\\[-1.8ex] & \% Alignment \\ 
\hline \\[-1.8ex] 
 Profession: Funeral Role & 0.121 (0.097) \\ 
  Profession: Medical Doctor & 0.028 (0.018) \\ 
  Profession: Nurse & 0.060 (0.058) \\ 
  Profession: Psychiatrist & 0.008 (0.017) \\ 
  Profession: Psychologist & $-$0.002 (0.008) \\ 
  Profession: Social Worker & 0.011$^{*}$ (0.006) \\ 
  Profession: Therapist & $-$0.002 (0.005) \\ 
  \textbf{is\_title\_visibleTrue} & \textbf{0.027$^{***}$ (0.009)} \\ 
  Subreddit: adultsurvivors & 0.045 (0.107) \\ 
  Subreddit: Advice & 0.204$^{*}$ (0.106) \\ 
  Subreddit: Anger & $-$0.042 (0.295) \\ 
  Subreddit: Anxiety & 0.193$^{*}$ (0.110) \\ 
  Subreddit: askatherapist & 0.105 (0.105) \\ 
  Subreddit: askfuneraldirectors & 0.012 (0.145) \\ 
  Subreddit: AskPsychiatry & 0.049 (0.106) \\ 
  Subreddit: bipolar & 0.123 (0.106) \\ 
  Subreddit: BipolarSOs & 0.391 (0.295) \\ 
  Subreddit: BodyAcceptance & 0.129 (0.191) \\ 
  Subreddit: BorderlinePDisorder & 0.029 (0.173) \\ 
  Subreddit: BPD & 0.141 (0.109) \\ 
  Subreddit: BPDlite & $-$0.525$^{*}$ (0.295) \\ 
  Subreddit: BPDlovedones & 0.099 (0.129) \\ 
  Subreddit: BreakUp & $-$0.059 (0.110) \\ 
  Subreddit: BreakUps & 0.174 (0.111) \\ 
  Subreddit: cancer & 0.075 (0.114) \\ 
  Subreddit: CaregiverSupport & 0.047 (0.123) \\ 
  Subreddit: CautiousBB & 0.053 (0.117) \\ 
  Subreddit: CBT & 0.145 (0.108) \\ 
  Subreddit: ChildrenofDeadParents & 0.201 (0.128) \\ 
  Subreddit: Codependency & 0.052 (0.125) \\ 
  Subreddit: CPTSD & 0.112 (0.107) \\ 
  Subreddit: cptsdcreatives & 0.475 (0.295) \\ 
  Subreddit: CPTSDNextSteps & 0.185 (0.191) \\ 
  Subreddit: datingoverfifty & 0.265$^{*}$ (0.144) \\ 
  Subreddit: datingoverforty & 0.206$^{*}$ (0.111) \\ 
  Subreddit: dbtselfhelp & 0.134 (0.173) \\ 
  \textbf{Subreddit: death} & \textbf{0.345$^{**}$ (0.162)} \\ 
  Subreddit: DecidingToBeBetter & 0.219$^{*}$ (0.114) \\ 
  Subreddit: dementia & 0.055 (0.136) \\ 
  Subreddit: depression & 0.157 (0.116) \\ 
  Subreddit: domesticviolence & 0.050 (0.128) \\ 
  Subreddit: emotionalabuse & $-$0.191 (0.191) \\ 
  Subreddit: emotionalneglect & 0.184$^{*}$ (0.110) \\ 
  Subreddit: ExNoContact & 0.026 (0.123) \\ 
  Subreddit: FriendsOver40 & 0.295 (0.221) \\ 
  \textbf{Subreddit: getdisciplined} & \textbf{0.307$^{**}$ (0.124)} \\ 
  Subreddit: grief & $-$0.088 (0.173) \\ 
  Subreddit: GriefSupport & 0.095 (0.108) \\
  \\
   \end{tabular}
  \begin{tabular}{@{\extracolsep{5pt}}lc} 
    \\[-1.8ex]\hline 
    \hline \\[-1.8ex] 
     & \multicolumn{1}{c}{\textit{Dependent variable:}} \\ 
    \cline{2-2} 
    \\[-1.8ex] & \% Alignment \\ 
    \hline \\[-1.8ex] 
  Subreddit: Grieving & 0.273 (0.295) \\ 
      
  Subreddit: happycryingdads & 0.475 (0.295) \\ 
   
  Subreddit: heartbreak & 0.473 (0.295) \\ 
  Subreddit: InternalFamilySystems & $-$0.038 (0.115) \\ 
  \textbf{Subreddit: IWantToLearn} & \textbf{0.314$^{***}$ (0.119)} \\ 
  Subreddit: lastimages & $-$0.155 (0.131) \\ 
  Subreddit: LifeAfterNarcissism & 0.113 (0.134) \\ 
  Subreddit: lonely & $-$0.038 (0.118) \\ 
  Subreddit: MadeMeCry & 0.475 (0.295) \\ 
  Subreddit: marriageadvice & 0.023 (0.221) \\ 
  Subreddit: mentalhealth & 0.068 (0.105) \\ 
  Subreddit: MentalHealthUK & 0.125 (0.107) \\ 
  Subreddit: Miscarriage & 0.076 (0.113) \\ 
  Subreddit: MomForAMinute & 0.066 (0.110) \\ 
  Subreddit: NarcAbuseAndDivorce & 0.098 (0.221) \\ 
  Subreddit: narcissism & 0.009 (0.191) \\ 
  Subreddit: NarcissisticAbuse & 0.089 (0.119) \\ 
  Subreddit: NarcissisticSpouses & $-$0.177 (0.191) \\ 
  Subreddit: OCD & 0.167 (0.110) \\ 
  Subreddit: offmychest & 0.068 (0.106) \\ 
  Subreddit: OldManDog & 0.173 (0.162) \\ 
  Subreddit: Petloss & 0.116 (0.148) \\ 
  Subreddit: PrayerRequests & 0.275 (0.221) \\ 
  Subreddit: PregnancyAfterLoss & 0.121 (0.108) \\ 
  Subreddit: productivity & 0.258$^{*}$ (0.154) \\ 
  \textbf{Subreddit: psychotherapy} & \textbf{0.234$^{**}$ (0.105)} \\ 
  Subreddit: sad & $-$0.192 (0.295) \\ 
  Subreddit: selfharm & 0.183 (0.148) \\ 
  Subreddit: selfimprovement & 0.090 (0.121) \\ 
  Subreddit: seniorkitties & 0.223$^{*}$ (0.121) \\ 
  Subreddit: SingleParents & 0.103 (0.107) \\ 
  \textbf{Subreddit: socialanxiety} & \textbf{0.368$^{**}$ (0.162)} \\ 
  \textbf{Subreddit: socialskills} & \textbf{0.284$^{**}$ (0.113)} \\ 
  Subreddit: SuicideBereavement & 0.041 (0.125) \\ 
  Subreddit: SuicideWatch & 0.016 (0.115) \\ 
  Subreddit: TalkTherapy & 0.191$^{*}$ (0.105) \\ 
  Subreddit: therapy & 0.118 (0.105) \\ 
  Subreddit: traumatoolbox & 0.137 (0.173) \\ 
  Subreddit: ttcafterloss & 0.096 (0.119) \\ 
  Subreddit: Vent & 0.069 (0.173) \\ 
  Subreddit: widowers & 0.275 (0.173) \\ 
  \textbf{Constant} & \textbf{0.527$^{***}$ (0.104)} \\ 
 \hline \\[-1.8ex] 
Observations & 20,029 \\ 
R$^{2}$ & 0.077 \\ 
Adjusted R$^{2}$ & 0.072 \\ 
Residual Std. Error & 0.276 (df = 19939) \\ 
F Statistic & 18.560$^{***}$ (df = 89; 19939) \\ 
\hline 
\hline \\[-1.8ex] 
\textit{Note:}  & \multicolumn{1}{r}{$^{*}$p$<$0.1; $^{**}$p$<$0.05; $^{***}$p$<$0.01} \\ 
\end{tabular} 
}
  \caption{Regression results on predicting the percentage of alignment with title (categorical variable), subreddit (categorical variable), and whether title is visible or not. Model coefficients ($\beta$) are shown for each factor and standard errors are shown in parentheses. Bolded rows show factors that are significant at at least p$<$0.05.} 
  \label{tab:is_title_visible} 
\end{table*}

\section{Subreddit-level Differences and Analysis}
\label{app:subreddit-analysis}

While we report aggregate statistics and trends, the subreddits in our studies still constitute distinct communities that vary in their behaviors (cf. Figures \ref{fig:target-pca} and \ref{fig:observer-pca}). Following, we report on two differences between subreddits that underscore themes in the main paper: the prevalence of giving Advice and the misalignments between Targets and Observers.

\subsection{Which subreddits give more Advice when asked?}

\paragraph{Setup}
We calculated the percentage of alignments where the Observer responded with Advice (i.e., the Target has requested advice), and averaged by subreddits.

\paragraph{Results}

\begin{figure}[t]
    \centering
    \includegraphics[width=0.47\textwidth]{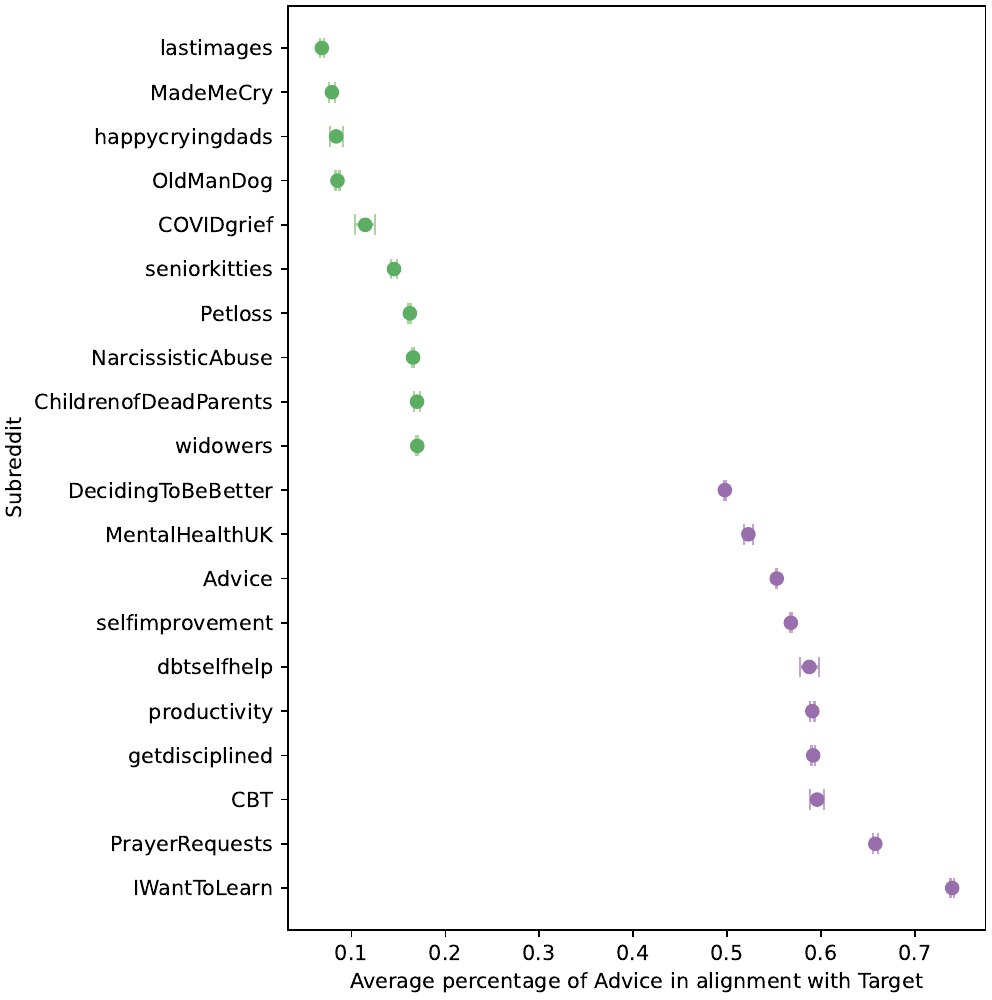}
    \caption{Percentage of alignments that Observer responding with Advice for each subreddit. Bottom-10 (least frequent) and top-10 (most frequent) are shown.}
    \label{fig:group_level_advice}
\end{figure}

While advice is present in all subreddits, as Figure \ref{fig:group_level_advice} shows, not all communities give advice.
Advice appears the most frequently when the Targets actively ask for it (cf. Figure \ref{fig:alignment}), yet in topics of loss and grief, Advice becomes much less frequent in the conversation despite the request for advice. However, at the other extreme, Targets in subreddits related to mental health receive an abundance of advice.
We hypothesize that in loss and grief subreddits, Observers instead focus on emotional support rather than suggestions, in part because of the difficulty of identifying what specifically can be done in such circumstances. 

\subsection{Misalignment for Appraisals}

Given the topical differences, do Observers in some subreddits align more closely with their Targets?

\paragraph{Setup}
For each appraisal/span in Target, we computed the percentage where the aligned appraisal in Observer is different from the Target and averaged by subreddits.

\paragraph{Results}

Table \ref{tab:appraisal_misalign} shows the most and least aligned subreddits for each appraisal type. Observers align well with targets around subreddits on loss and grief in Pleasantness, Certainty, and Objective Experience. 
Anticipated Effort and Situational Control are mainly aligned best with topics around mental health. Noticeably, Self-other Agency has a bias towards alignment for abuse-related topics. As confirmed with what we have observed, Advice appears in observers the most when it's actively been asked for. 

\camerareadytext{
Less clustered than most aligned topics, appraisals could be misaligned in a diverse range of subreddits. However, the exception appears for Self-other agency, which observers seldom correctly align with targets in mental health topics.
}

\begin{table*}[!b]
\resizebox{\textwidth}{!}{
\begin{tabular}{rcl}
\textbf{Appraisal/Span} & \multicolumn{2}{c}{\textbf{Subreddits}} \\ \hline
 & Most misaligned & \begin{tabular}[c]{@{}l@{}}selfimprovement, getdisciplined, productivity, socialskills, askfuneraldirectors, \\ CBT, Advice, PrayerRequests, IWantToLearn, AskPsychiatry\end{tabular} \\
\multirow{-2}{*}{\textbf{Pleasantness}} & \cellcolor[HTML]{EFEFEF}Least misaligned & \cellcolor[HTML]{EFEFEF}\begin{tabular}[c]{@{}l@{}}MadeMeCry, COVIDgrief, lastimages, Petloss, Miscarriage, GriefSupport, \\ Grieving, widowers, happycryingdads, grief\end{tabular} \\
 & Most misaligned & \begin{tabular}[c]{@{}l@{}}gaslighting, sad, happycryingdads, marriageadvice, narcissism, IWantToLearn, \\ askatherapist, askfuneraldirectors, Advice, AskPsychiatry\end{tabular} \\
\multirow{-2}{*}{\textbf{Anticipated Effort}} & \cellcolor[HTML]{EFEFEF}Least misaligned & \cellcolor[HTML]{EFEFEF}\begin{tabular}[c]{@{}l@{}}CPTSDNextSteps, ttcafterloss, FriendsOver40, widowers, PregnancyAfterLoss,\\  BPDlite, Miscarriage, cptsdcreatives, CPTSD, MadeMeCry\end{tabular} \\
 & Most misaligned & \begin{tabular}[c]{@{}l@{}}dbtselfhelp, MomForAMinute, NarcAbuseAndDivorce, psychotherapy, IWantToLearn, \\ BodyAcceptance, marriageadvice, Advice, askfuneraldirectors, PrayerRequests\end{tabular} \\
\multirow{-2}{*}{\textbf{Situational Control}} & \cellcolor[HTML]{EFEFEF}Least misaligned & \cellcolor[HTML]{EFEFEF}\begin{tabular}[c]{@{}l@{}}BPDlite, happycryingdads, CPTSD, widowers, Anxiety, BPD, ChildrenofDeadParents,\\  depression, COVIDgrief, SuicideBereavement\end{tabular} \\
 & Most misaligned & \begin{tabular}[c]{@{}l@{}}OCD, AskPsychiatry, askfuneraldirectors, MentalHealthUK, dbtselfhelp, \\ PrayerRequests, IWantToLearn, getdisciplined, CautiousBB, productivity\end{tabular} \\
\multirow{-2}{*}{\textbf{Self-other Agency}} & \cellcolor[HTML]{EFEFEF}Least misaligned & \cellcolor[HTML]{EFEFEF}\begin{tabular}[c]{@{}l@{}}NarcissisticAbuse, gaslighting, NarcissisticSpouses, emotionalabuse, BPDlovedones,\\  abusiverelationships, marriageadvice, BreakUp, EmotionalAbuseSupport, ExNoContact\end{tabular} \\
 & Most misaligned & \begin{tabular}[c]{@{}l@{}}DecidingToBeBetter, selfimprovement, CancerCaregivers, happycryingdads, Anger,\\  dbtselfhelp, getdisciplined, MentalHealthUK, productivity, IWantToLearn\end{tabular} \\
\multirow{-2}{*}{\textbf{Certainty}} & \cellcolor[HTML]{EFEFEF}Least misaligned & \cellcolor[HTML]{EFEFEF}\begin{tabular}[c]{@{}l@{}}death, OldManDog, lastimages, Petloss, heartbreak, grief,\\  ChildrenofDeadParents, widowers, BreakUps, gaslighting\end{tabular} \\
 & Most misaligned & \begin{tabular}[c]{@{}l@{}}CBT, Anger, abusiverelationships, marriageadvice, domesticviolence, \\ EmotionalAbuseSupport, Advice, emotionalabuse, gaslighting, PrayerRequests\end{tabular} \\
\multirow{-2}{*}{\textbf{Objective Experience}} & \cellcolor[HTML]{EFEFEF}Least misaligned & \cellcolor[HTML]{EFEFEF}\begin{tabular}[c]{@{}l@{}}CautiousBB, ttcafterloss, PregnancyAfterLoss, Miscarriage, COVIDgrief, \\ ChildrenofDeadParents, lastimages, cancer, GriefSupport, Grieving\end{tabular} \\
 & Most misaligned & \begin{tabular}[c]{@{}l@{}}CPTSDNextSteps, BorderlinePDisorder, BPD4BPD, widowers, NarcissisticAbuse,\\  BPD, emotionalneglect, cptsdcreatives, BPDlite, CPTSD\end{tabular} \\
\multirow{-2}{*}{\textbf{Advice}} & \cellcolor[HTML]{EFEFEF}Least misaligned & \cellcolor[HTML]{EFEFEF}\begin{tabular}[c]{@{}l@{}}IWantToLearn, Advice, MentalHealthUK, PrayerRequests, CancerCaregivers,\\  AskPsychiatry, CBT, MomForAMinute, selfimprovement, MMFB\end{tabular}
\end{tabular}
}
\caption{Ten most misaligned and ten least misaligned subreddits for each appraisal/span.}
\label{tab:appraisal_misalign}
\end{table*}

\section{Model output examples on alignment prediction: qualitative error analysis}

Table \ref{tab:align_wrong_examples} shows examples of positive and negative classification errors for alignment prediction, along with descriptions of what pattern was seen for this type of error.

\begin{table*}[tb]
\resizebox{\textwidth}{!}{
\begin{tabular}{clcc}
 & \multicolumn{1}{c}{} & \multicolumn{2}{c}{\textbf{Aligned?}} \\
\multirow{-2}{*}{\textbf{Desciption}} & \multicolumn{1}{c}{\multirow{-2}{*}{\textbf{Example}}} & \textbf{Model} & \textbf{Label} \\ \hline
Overgeneralization & \begin{tabular}[c]{@{}l@{}}\textbf{Target:} IDK, I guess despair is just all over in me.\\ \textbf{Observer:} When things go so badly that you want to see it become even worse.\end{tabular} & Yes & No \\
\rowcolor[HTML]{EFEFEF} 
\begin{tabular}[c]{@{}c@{}}Pattern-overcatching\\ (First, Second, ...)\end{tabular} & \begin{tabular}[c]{@{}l@{}}\textbf{Target:} I don't really know what to do.\\ \textbf{Observer:} Second, any thoughts you're having are nothing to feel guilty about \\ or be as homes of.\end{tabular} & Yes & No \\
\begin{tabular}[c]{@{}c@{}}Wrong object\\ ("he" and "I")\end{tabular} & \begin{tabular}[c]{@{}l@{}}\textbf{Target:} Am I in the wrong? Do I need therapy to help me get over his past \\ hurtful behavior...\\ \textbf{Observer:} You aren't obligated to feel or act a certain way to make him feel \\ connected to you or lessen any guilt he may feel.\end{tabular} & Yes & No \\
\rowcolor[HTML]{EFEFEF} 
Implicit reference & \begin{tabular}[c]{@{}l@{}}\textbf{Target:} I feel like I'm alone all the time so I might as well just be alone\\ \textbf{Observer:} I feel like you are so occupied with what you don't have, you're \\ thinking about what you donn't have.\end{tabular} & No & Yes \\
Explicit reference & \begin{tabular}[c]{@{}l@{}}\textbf{Target:}  But why does it still hurt..\\ \textbf{Observer:} It takes a long time sometimes to get over someone.\end{tabular} & No & Yes \\
\rowcolor[HTML]{EFEFEF} 
Valid alignment in experience & \begin{tabular}[c]{@{}l@{}}\textbf{Target:} She broke up with me, she destroyed my heart and she's the one \\ finding someone new to be happy with while i haven't had even a crush \\ on anyone in all this time.\\ \textbf{Observer:} i also felt like i was loosing in the breakup because i could \\ not move on as fast\end{tabular} & No & Yes
\end{tabular}
}
\caption{Examples of errors that our best model makes when predicting alignment.}
\label{tab:align_wrong_examples}
\end{table*}

\end{document}